\documentclass[times,preprint,authoryear,10pt]{elsarticle}

\usepackage{booktabs}
\usepackage{amssymb}
\usepackage{rotating}
\usepackage{amsmath,amsfonts,bm}

\def\eqref#1{equation~\ref{#1}}
\def\Eqref#1{Equation~\ref{#1}}
\def\1{\bm{1}}

\DeclareMathAlphabet{\mathsfit}{\encodingdefault}{\sfdefault}{m}{sl}
\SetMathAlphabet{\mathsfit}{bold}{\encodingdefault}{\sfdefault}{bx}{n}

\newcommand{\ablCostOpen}{0.3511}

\newcommand{\ablRatioClosed}{1.2}

\newcommand{\ablRatioOpen}{8.3}

\newcommand{\accEasyLocal}{0.723}
\newcommand{\accEasyLocalCaught}{129}
\newcommand{\accEasyPooled}{0.963}
\newcommand{\accHardLocal}{0.519}
\newcommand{\accHardPooled}{0.727}
\newcommand{\boundAllAbove}{11}
\newcommand{\boundAllAboveMax}{0.0091}
\newcommand{\boundAllCells}{35}
\newcommand{\boundAllMean}{0.004}
\newcommand{\boundExcessFold}{46}
\newcommand{\boundGlobalHi}{98}
\newcommand{\boundGlobalLo}{70}
\newcommand{\boundGroupExcess}{0.42}
\newcommand{\boundLocalHi}{99}
\newcommand{\boundLocalLo}{98}
\newcommand{\boundLocalRawMax}{0.0010}
\newcommand{\boundNDensities}{7}
\newcommand{\capacityFold}{eightfold}
\newcommand{\closedErr}{0.0011}
\newcommand{\closedFar}{0.965}
\newcommand{\closedNine}{0.5478}
\newcommand{\closedRmax}{513}
\newcommand{\cutAblHi}{0.79}
\newcommand{\cutAblLo}{0.28}
\newcommand{\cutAblRet}{0.527}
\newcommand{\cutAblSD}{0.157}
\newcommand{\cutNoiseFold}{8}
\newcommand{\cutRetHi}{0.97}
\newcommand{\cutRetLo}{0.90}
\newcommand{\cutRetRet}{0.925}
\newcommand{\cutRetSD}{0.019}
\newcommand{\deffLevels}{8}
\newcommand{\deffMarginMax}{0.9}
\newcommand{\deffNullEasy}{46.2}
\newcommand{\deffNullHard}{23.0}
\newcommand{\deffObsEasy}{33.7}
\newcommand{\deffObsHard}{25.5}
\newcommand{\deffObsPeak}{40.0}

\newcommand{\derivAtTenN}{9.9}
\newcommand{\derivAtTenObs}{0.1290}
\newcommand{\derivAtTenPred}{0.1269}
\newcommand{\derivConst}{0.399}
\newcommand{\derivMaxErr}{4}
\newcommand{\derivNMin}{3.0}
\newcommand{\disBandFold}{eight}

\newcommand{\disClosedHi}{1.3}
\newcommand{\disClosedLo}{1.1}

\newcommand{\disOpenHi}{9.8}
\newcommand{\disOpenLo}{9.1}

\newcommand{\evalSeq}{512}

\newcommand{\factFstFold}{25}
\newcommand{\factFstHi}{0.24}
\newcommand{\factFstLo}{0.010}
\newcommand{\factLevels}{5}
\newcommand{\factRatioHi}{8.3}
\newcommand{\factRatioLo}{6.3}
\newcommand{\genDoseN}{0.29, 0.99, 3.00 and 9.89}
\newcommand{\genDoseRatio}{8.3, 3.1, 1.7 and 1.3}
\newcommand{\genHiAcc}{0.951}
\newcommand{\genLoAcc}{0.577}

\newcommand{\idealCFold}{2.2}

\newcommand{\introFull}{0.976}
\newcommand{\introLocal}{0.548}
\newcommand{\introNet}{0.967}
\newcommand{\introOracle}{0.976}
\newcommand{\jacC}{4}
\newcommand{\jacClnPast}{0.3164}
\newcommand{\jacDiag}{0.9766}
\newcommand{\jacErr}{5.6\times 10^{-16}}
\newcommand{\jacInvS}{0.0039}
\newcommand{\jacL}{64}
\newcommand{\jacOffdiag}{0.0046}
\newcommand{\jacRowTotal}{1\times 10^{-16}}
\newcommand{\jacRowsum}{-0.9766}
\newcommand{\jacS}{256}
\newcommand{\lawAlphaGen}{0.94}
\newcommand{\lawAlphaGenSE}{0.04}
\newcommand{\lawAlphaSyn}{1.06}

\newcommand{\lawAlphaSynSE}{0.04}
\newcommand{\lawC}{3.94}

\newcommand{\lawRootFold}{23}

\newcommand{\lawSpread}{22}

\newcommand{\matchedGenRatios}{8.3$\times$, 3.1$\times$, 1.7$\times$, 1.3$\times$}
\newcommand{\matchedSynN}{0.27, 0.97, 2.97, 9.74}
\newcommand{\matchedSynRatios}{15.6$\times$, 4.3$\times$, 2.1$\times$, 1.4$\times$}
\newcommand{\nParams}{226{,}177}
\newcommand{\nParamsLo}{26{,}497}
\newcommand{\nSeeds}{three}
\newcommand{\normBatch}{+0.3687}
\newcommand{\normBatchSD}{0.0046}
\newcommand{\normFst}{0.0380}
\newcommand{\normGroup}{+0.0452}
\newcommand{\normGroupSD}{0.0107}
\newcommand{\normInst}{+0.0424}
\newcommand{\normInstSD}{0.0080}
\newcommand{\normNone}{+0.3740}
\newcommand{\normNoneSD}{0.0052}
\newcommand{\normPos}{+0.3741}
\newcommand{\normPosSD}{0.0052}
\newcommand{\numExperiments}{729}
\newcommand{\optimaCross}{0.0001}
\newcommand{\optimaCrossN}{0.09}
\newcommand{\paramFold}{8.5}
\newcommand{\perPosBayes}{0.5160}
\newcommand{\portFstLo}{0.0030}
\newcommand{\portLevels}{8}
\newcommand{\portMaxDiff}{0.0029}
\newcommand{\portMaxRel}{14}
\newcommand{\reachBracketHi}{33}
\newcommand{\reachFold}{227}
\newcommand{\reachMax}{2049}
\newcommand{\reachMin}{9}
\newcommand{\realAbl}{0.2919}
\newcommand{\realAblRatio}{16.1}
\newcommand{\realAblSD}{0.0701}
\newcommand{\realFst}{0.0322}
\newcommand{\realGroup}{+0.0181}
\newcommand{\realGroupSD}{0.0105}
\newcommand{\realHiAcc}{0.8717}
\newcommand{\realLoAcc}{0.8536}
\newcommand{\realNoiseFold}{6.7}
\newcommand{\realPooledEndHi}{+0.020}
\newcommand{\realPooledEndLo}{+0.015}
\newcommand{\realPos}{+0.2190}
\newcommand{\realPosEndFirst}{+0.030}
\newcommand{\realPosEndLast}{+0.219}
\newcommand{\realPosSD}{0.0195}
\newcommand{\realRatio}{12.1}

\newcommand{\realRatioSeedHi}{19}
\newcommand{\realRatioSeedLo}{8}
\newcommand{\realRet}{0.0181}
\newcommand{\realRetSD}{0.0105}

\newcommand{\realSwitch}{0.13}

\newcommand{\retCostOpen}{0.0423}

\newcommand{\shiftChanged}{8.9}
\newcommand{\shiftSingle}{74}

\newcommand{\shortSparse}{0.009}
\newcommand{\shortSparseFull}{0.009}
\newcommand{\shortWorst}{0.037}
\newcommand{\sigCells}{21}
\newcommand{\sigCorrHi}{0.92}
\newcommand{\sigCorrLo}{0.50}

\newcommand{\sigRatioHi}{1.02}
\newcommand{\sigRatioLo}{0.99}
\newcommand{\surveyDomainList}{action segmentation, genomics, sleep staging and speaker diarization}
\newcommand{\surveyDomains}{four}
\newcommand{\surveyExposed}{Two}

\newcommand{\surveyN}{thirteen}
\newcommand{\surveyUnexposed}{eleven}
\newcommand{\sweepGroupHi}{+0.0524}
\newcommand{\sweepGroupLo}{+0.0320}
\newcommand{\sweepPosHi}{+0.3741}
\newcommand{\sweepPosLo}{+0.2011}
\newcommand{\synDelta}{0.08}
\newcommand{\synDoseFold}{212}

\newcommand{\synNMax}{19.9}
\newcommand{\synNMin}{0.094}
\newcommand{\synRatioMatched}{15.6}
\newcommand{\synRatioMax}{62.8}
\newcommand{\synRatioMin}{1.2}
\newcommand{\synSpanRatio}{329}
\newcommand{\tasCeiling}{0.9522}

\newcommand{\tasClnRatio}{6.5}

\newcommand{\tasGln}{+0.0161}
\newcommand{\tasGlnLo}{0.9246}
\newcommand{\tasGlnRatio}{20.6}
\newcommand{\tasGlnSD}{0.0059}
\newcommand{\tasParams}{226{,}538}
\newcommand{\tasPos}{+0.3322}
\newcommand{\tasPosSD}{0.0030}
\newcommand{\tasRatioSeedHi}{32}
\newcommand{\tasRatioSeedLo}{15}
\newcommand{\tasReachHi}{1027}
\newcommand{\tasReachLo}{7}
\newcommand{\tasSwitch}{0.27}
\newcommand{\trainSeq}{3840}

\newcommand{\unetDepths}{6}
\newcommand{\unetFlat}{0.017}

\newcommand{\unetMatchDepth}{4}
\newcommand{\unetMatchParams}{220{,}609}
\newcommand{\unetMatchPct}{2.5}

\newcommand{\unetParamsHi}{319{,}809}
\newcommand{\unetParamsLo}{71{,}809}
\newcommand{\unetPctOracle}{98.9}
\newcommand{\unetPoolFold}{32}

\newcommand{\unetRFFold}{35}
\newcommand{\unetRFHi}{768}
\newcommand{\unetRFLo}{22}
\newcommand{\unetRatioDense}{2.0}
\newcommand{\unetRatioSparse}{16.1}

\newcommand{\unetSwDense}{2.98}
\newcommand{\unetSwSparse}{0.27}
\newcommand{\unetValuesDeep}{4{,}096}
\newcommand{\unetWindow}{4096}
\newcommand{\window}{4096}
\newcommand{\xfDensities}{four}
\newcommand{\xfDiffHi}{0.0025}
\newcommand{\xfDiffLo}{0.0021}
\newcommand{\xfGap}{0.0047}

\usepackage{microtype}
\usepackage{etoolbox}
\AtBeginEnvironment{tabular}{\small}
\usepackage[hidelinks]{hyperref}

\journal{}

\begin{document}

\begin{frontmatter}

% Running (short) title, for the submission form; elsarticle prints no
% running head itself, Elsevier production applies it:
%   Normalization can supply a sequence labeler's context
\title{Beyond receptive fields: sequence-pooled normalization can supply most of a sequence labeler's context}

% Neural Networks reviews single-blind, so the author block is named. Replace
% with an anonymous block if a double-anonymized submission is chosen instead.
\author{Qing Tian}
\address{Department of Computer Science, University of Alabama at Birmingham, Birmingham, AL 35294, USA}
\ead{qtian@uab.edu}

\begin{abstract} A convolutional sequence labeler's receptive field is routinely treated as the extent of the model's usable context: it sets dilation schedules, bounds streaming horizons, and underwrites locality claims. However, we show that this can be false: when a normalization layer computes statistics from the current input along the sequence at inference, those statistics open a sequence-spanning path that bypasses the convolutional receptive field to provide global context. We derive this from the layer's Jacobian (the criterion needs no experiment), and what the path carries has a closed form. On a synthetic labeling process with computable optima, the global summary that a sequence-spanning normalization encodes already supplies almost all of what a larger receptive field would buy where labels come in long runs: a network reaching \reachMin\ positions comes within \shortSparseFull\ of the whole-sequence optimum, against a near-chance bound for its reach. Closing the path, by taking the same statistics per position, multiplies what enlarging the receptive field is worth by up to an order of magnitude on simulated genomes at every difficulty level tested and on real \mbox{1000 Genomes} haplotypes. The same path also confounds attribution: ablating a trained network's receptive-field-enlarging blocks severs part of the path, overstating their contribution \ablRatioOpen--\realAblRatio-fold relative to retraining from scratch. The substitution of normalization for receptive field fades as labels switch more often. Where labels run long, neither the receptive-field justification nor the ablation is wrong about its numbers, but both credit the wrong component. \end{abstract}


\end{frontmatter}

\section{Introduction}
\label{sec:intro}

\begin{figure}[htbp]
\centering
\includegraphics[width=0.98\linewidth]{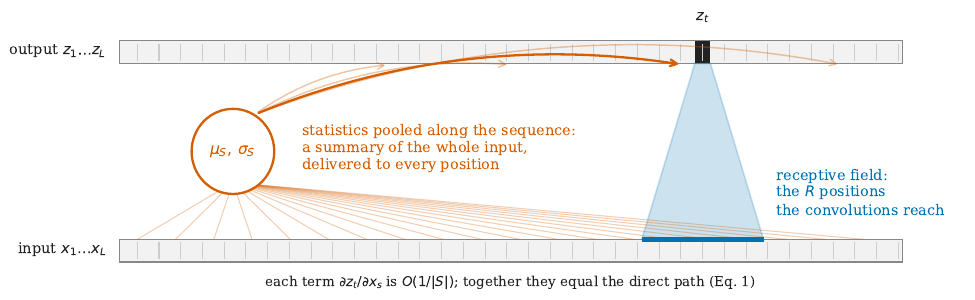}
\caption{The two paths from input to output. Blue: the receptive field, the bounded set of positions the convolutions reach, which is the quantity architectures report and justify. Orange: the path a normalization layer opens when its statistics $\mu_S, \sigma_S$ are pooled along the sequence: every position contributes to the statistic, and the statistic reaches every position. Each single term $\partial z_t/\partial x_s$ is $O(1/|S|)$ (with $|S|$ the number of positions pooled), small enough to dismiss, but the terms do not cancel one another: they sum to exactly minus the position's own term $\partial z_t/\partial x_t$, so the normalization path is as large as the convolutional one it shadows (Section~\ref{sec:theory}). No receptive-field calculation accounts for it, because it passes through no convolution.}
\label{fig:mech}
\end{figure}

An architecture's receptive field, or reach,\footnote{Throughout, the receptive field means the span the convolutions reach, computed from kernel sizes, dilations and strides. Section~\ref{sec:bounds} measures the set of inputs that actually influences an output, and the two agree unless a normalization statistic spans the sequence.} is a design parameter and a justification. Dilated stacks are specified by the span they achieve, streaming systems budget their latency against it, and when the contribution of that span is questioned, the standard answer removes the reach-enlarging blocks from a trained network and reports what is lost. All three practices treat the receptive field as the extent of the context available to the model. This paper shows that the description fails whenever the network contains normalization whose statistics are computed along the sequence at inference, and it measures by how much. Such a layer hands every position a whole-input summary, through a path no receptive-field calculation accounts for, because it passes through no convolution (Figure~\ref{fig:mech}). We find that where labels come in long runs, that single summary supplies almost all of the usable context, and both the receptive-field justification and the ablation credit that context to the wrong component.

\begin{figure}[htbp]
\centering
\includegraphics[width=0.98\linewidth]{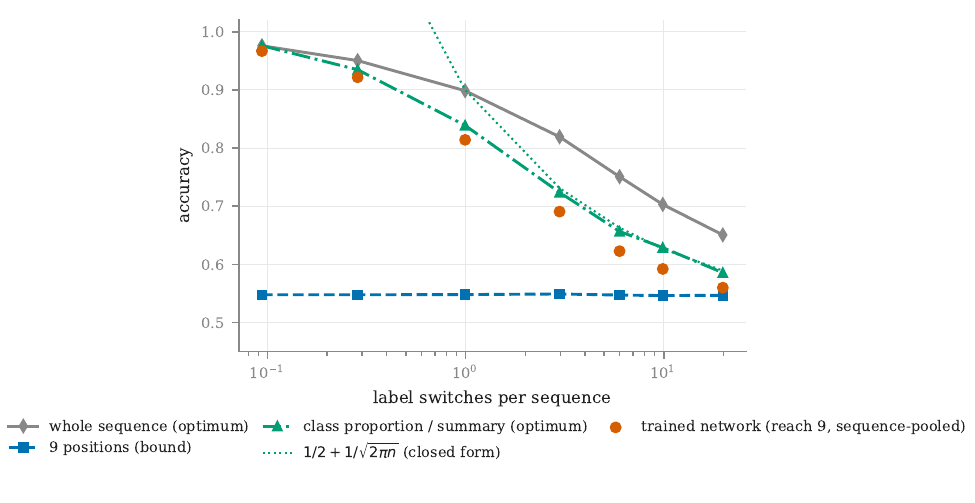}
\caption{The information landscape. The horizontal axis is $n$, the expected number of label switches in a sequence, so $n$ below one means the most likely number of switches in a sequence is zero: one label from end to end. Three of the four lines are references, properties of the generating process computed before any model was trained, with nothing fitted: the whole-sequence optimum (gray, solid), the class-proportion oracle (green, dash-dotted), and a conservative bound for \reachMin\ positions of context (blue, dashed). The green class-proportion oracle is the summary a sequence-spanning normalization encodes, to first order. The fourth, the dotted curve, is the closed form derived in Section~\ref{sec:theory}, also with nothing fitted, which tracks the proportion oracle for $n \gtrsim 1$ and necessarily departs below it. Orange points are trained networks (each for one switch rate): the dilated residual stack of \ref{app:arch} cut to its first block (receptive field \reachMin\ positions), with normalization statistics pooled along the whole sequence rather than taken at each position. The networks track the proportion curve rather than the bound for their own receptive field, falling short of the proportion oracle by \shortSparse\ where labels rarely switch and by at most \shortWorst\ anywhere.}
\label{fig:decomp}
\end{figure}

To say how much such a summary supplies, we need an absolute scale. On sequence-labeling processes whose generating distribution is known, three reference quantities can be computed rather than estimated from trained models: a conservative bound on the best accuracy available from a bounded span of positions, the best available from the whole sequence, and the best available from the sequence's class proportion alone, a single number that a sequence-spanning normalization statistic encodes to first order (Section~\ref{sec:theory}). Placing trained networks against those three references puts their accuracies on that scale. Figure~\ref{fig:decomp} does this for a synthetic process with computable optima. Where labels rarely switch (left end in Figure~\ref{fig:decomp}), the references line up: an oracle handed only the class proportion reaches \introOracle\ and perfect use of all \window\ positions gives \introFull, against \introLocal\ for the bound at \reachMin\ positions of reach. A network with sequence-pooled normalization reaching only \reachMin\ positions attains \introNet: only \shortSparseFull\ below the whole-sequence optimum, far above what its receptive field reaches, and tracking the proportion oracle. In this regime, the path is not a small correction to the receptive field but the dominant source of context.

The same path confounds attribution. Ablation removes a component from a trained network and re-evaluates.\footnote{Both this and training the reduced architecture from scratch can be called ablation in practice, the first in interpretability work and the second in the ablation tables of architecture papers. This paper compares them, so it calls the second retraining (Section~\ref{sec:ablation}).} Applied to the blocks that enlarge the receptive field, it removes their normalization layers with them, severing part of the sequence-spanning path and charging the blocks for context that a network trained from scratch without them obtains anyway. On the simulated genomic task, the ablated cost exceeds the retrained one \ablRatioOpen-fold, and \realAblRatio-fold on real \mbox{1000 Genomes} haplotypes. Where no statistic spans the sequence, ablation and retraining nearly agree (differing by no more than the adaptation and compensation that follow any removal \citep{mcgrath2023hydra}), with everything else held fixed (Section~\ref{sec:ablation}). The discrepancy is therefore a function of the normalization's pooling axes alone.

The effect is conditional, and its two conditions can be checked before any experiment: normalization statistics computed from the current input, along the sequence, at inference, which follows from the layer's definition; and labels that come in long runs, which follows from the task. Where either fails, the account predicts no effect. Section~\ref{sec:tests} names two cases of predicted absence in advance and finds the effect absent in both. Where both hold, the failure is not the models', which are doing something sensible with the information available to them. It is a failure of the description, and it has practical cost wherever the description is acted on: an architect sizing dilation schedules by measured gains, a streaming system relying on a bounded horizon it has by construction and not in fact \citep{luo2019convtasnet}, a planner relying on locality for temporal compositionality \citep{janner2022planning}.

Normalization carrying information past the receptive field has been observed before: \citet{pfrommer2025spooky} show it on a synthetic localization task, via iterative message passing between positions whose receptive fields overlap. We arrived at the phenomenon independently and by a different mechanism, the broadcast they set aside: a single layer's pooled statistic needs neither depth nor overlap. The rest of what is new lies downstream of existence, which their task, admitting no competing source of evidence, could not measure: the derived criterion, a closed form for what an open path carries, the share it carries in practice when a real input--label relationship competes with it, and the consequence for attribution.
The paper's contributions group under three headings:
\begin{itemize}
\item \textbf{Mechanism.} A criterion derived from a normalization layer's Jacobian that determines, from the layer's definition alone, whether it opens a sequence-spanning path; and a closed form for what such a path carries ($1/\sqrt{2\pi n}$ above chance, where $n$ is the expected number of label switches in a sequence), confirmed against exactly computed optima to within \derivMaxErr\% with nothing fitted (Section~\ref{sec:theory}). A survey of \surveyN\ published models shows the criterion partitions real architectures, and that the partition follows neither the domain nor the reported receptive field (Section~\ref{sec:discussion}).
\item \textbf{Magnitude.} Four predictions of that account, tested across two unrelated generating processes, real \mbox{1000 Genomes} haplotypes, a published Conv-TasNet separator, and a transformer whose attention already spans the sequence, and each supported: reach is nearly worthless while the path is open; only pooling along the sequence at inference produces the effect; the effect decays as labels switch more often; and it is absent in the two cases where the account predicts no effect. Two further checks show the networks sit at the computed bounds, and that the account survives a U-Net whose pooled extent shrinks with depth (Section~\ref{sec:tests}).
\item \textbf{Consequence.} Where the removed blocks carry sequence-spanning normalization, block ablation severs part of the path with them and charges them for it, overstating their contribution \ablRatioOpen-fold in simulation and \realAblRatio-fold on real haplotypes. With only per-position statistics, the overstatement falls to the level retraining alone produces, isolating the path as the cause (Section~\ref{sec:ablation}).
\end{itemize}

\section{Related work}

\subsection{Normalization as an information path} That normalization statistics pooled over spatial extent are global summaries is not new; it is the operating premise of instance normalization in style transfer, where the spatial mean and variance of a feature map are treated as global style and transplanted between images \citep{ulyanov2016instance, huang2017adain}. The same property is familiar in a different guise as leakage along the batch axis, which is why batch statistics are replaced by running averages at evaluation \citep{ioffe2015batch}, a replacement that itself costs accuracy at small batch size, where the two sets of statistics diverge \citep{wu2021rethinking}. \citet{pfrommer2025spooky} draw out the consequence for receptive field, and additionally observe that BatchNorm's path vanishes once population statistics replace minibatch ones at evaluation, a behavior the criterion of Section~\ref{sec:theory} predicts from the layer's definition and Section~\ref{sec:tests} confirms. Three things in this paper appear in none of the work above: the criterion, which decides from a layer's definition alone which layers open a path; \Eqref{eq:advantage}, which gives what the path carries in closed form, against exactly computed optima rather than against a trained baseline; and the survey of Section~\ref{sec:discussion}, which applies the criterion to each model's released code, which is what a reader can repeat on their own model.

Normalization statistics computed at inference, rather than frozen after training, are transductive: the prediction depends on data present alongside the point being predicted. That character is also exploited deliberately. Test-time adaptation recomputes normalization statistics on the test data and gains robustness by doing so \citep{schneider2020improving, wang2021tent}, treating the statistic as a path from the test distribution to the prediction. The path studied here runs from distant positions within the current input to the prediction, and the issue is not that such a path exists but that the quantity practitioners use to describe a model's context does not account for it. A separate line establishes that convolutional networks acquire positional information through zero-padding at the boundaries \citep{islam2020position, kayhan2020translation}; that path has a distinct signature, strongest at the edges of the input and decaying inward, while the effect studied here is uniform across the sequence.

\subsection{Attributing function to components} Removing a component and re-evaluating is the standard method for attributing function, and its limitations are documented. Redundancy makes single-unit ablation uninformative \citep{morcos2018single, meyes2019ablation}; in residual and transformer architectures the remaining components compensate, so the loss understates what the component does \citep{mcgrath2023hydra}; and the choice of what to substitute for the removed component changes the answer \citep{li2024optimal}. Whether the model is re-evaluated after removal or retrained without the component changes the conclusion outright: retraining after removing the features that an attribution method identifies overturns the ranking of those methods \citep{hooker2019benchmark}, and in structured pruning, architectures retrained from scratch can match their pruned-and-fine-tuned counterparts \citep{liu2019rethinking}. Those concern the surviving components, or what is substituted for the removed one, and they predict a discrepancy between removal and retraining in general. The confound we report is different in kind: ablation removes a shared path along with the component, and charges the component for both. Its signature distinguishes it, because the size of the discrepancy is a function of the normalization scheme. Where statistics pooled along the sequence open a path, ablation and retraining differ by roughly an order of magnitude. Where they do not, the two differ only by the factor adaptation and compensation account for, with every other aspect of the experiment unchanged (Section~\ref{sec:ablation}).

\subsection{Receptive field as justification} Reach is routinely the stated reason for an architecture, in more fields than one. Dilated stacks are motivated by the span they achieve \citep{yu2016dilated, oord2016wavenet, bai2018tcn}. Temporal action segmentation is largely built on them and reports the receptive field as what lets a model see a whole action \citep{farha2019mstcn, yi2021asformer}; sleep staging quotes it in minutes of signal \citep{perslev2021usleep}; speaker diarization inherits that span from the convolutional frontends beneath its segmentation models \citep{bredin2020pyannote}; and \citet{luo2019convtasnet} report the receptive field of the Conv-TasNet separator as a design parameter. Genomic sequence models are compared on the distance they integrate, from the 131\,kb input of Basenji \citep{kelley2018sequential} to the 197\,kb of Enformer \citep{avsec2021effective}, justified by the regulatory elements it brings within reach. \citet{luo2016effective} note that the nominal receptive field overstates what a network uses, since the gradient-weighted extent is smaller; we report the opposite failure, in which the nominal field understates what the network can access.

\section{Theory: what context is available, and from where}
\label{sec:theory}

The claim that a receptive field understates a model's context turns on two questions. Whether a layer opens a path at all, meaning that distant inputs influence an output through it, is a question of dependency and a property of the layer, settled by differentiating it, without running an experiment or specifying a task. What the path carries is a question of information and a property of the task, and needs the task's generating process. Separating the two is what lets the answer be conditional and still definite; this section takes them in that order.

\subsection{What the path is} A normalization layer standardizes each entry against statistics pooled over an index set $S$, writing $z_t = \gamma\,(x_t - \mu_S)/\sigma_S + \beta$ with $\mu_S = |S|^{-1}\sum_{s \in S} x_s$ and $\sigma_S^2 = |S|^{-1}\sum_{s \in S} (x_s - \mu_S)^2$, the population convention the implementations use. Differentiating gives, for every $s \in S$,

\begin{equation}
\frac{\partial z_t}{\partial x_s}
  \;=\; \frac{\gamma}{\sigma_S}
  \Big[\,\delta_{ts} - \frac{1 + \hat x_t \hat x_s}{|S|}\,\Big],
  \qquad \hat x = (x - \mu_S)/\sigma_S ,
\label{eq:jacobian}
\end{equation}
which we checked against automatic differentiation to within $10^{-15}$ (\ref{app:bounds}).\footnote{Implementations normalize by $\sqrt{\sigma_S^2+\epsilon}$; \Eqref{eq:jacobian} is exact at $\epsilon = 0$, which is how the check is run. For $\epsilon > 0$ the same expression holds after replacing $\sigma_S$ by $\sqrt{\sigma_S^2+\epsilon}$ in both the prefactor and the definition of $\hat x$; adding a constant to every $x_s$ still leaves $z_t$ unchanged, so the sum of \Eqref{eq:jacobian} over $s$ is still exactly zero and the $s \neq t$ terms generically remain nonzero.} Three consequences follow, settling when the path exists, why one-position-at-a-time analysis misses it, and which layers have it.

\begin{enumerate}
\item \textbf{The path exists whenever $S$ spans the current input's sequence at inference.} The off-diagonal terms, $\partial z_t/\partial x_s$ with $s \neq t$, are generically nonzero, whatever the convolutions reach. No receptive-field calculation based only on the convolutions accounts for them, because none of them passes through a convolution. This is a property of the layer, and downstream computation could in principle cancel it at the output. Whether it does is measured rather than assumed, since the receptive-field measurement of Section~\ref{sec:bounds} differentiates the whole network and finds the whole sequence present.

\item \textbf{It is invisible one position at a time.} Each off-diagonal term is $O(1/|S|)$: for the GroupNorm of Section~\ref{sec:setup}, whose groups span $8$ channels and \window\ positions, about $3\times10^{-5}$ of the diagonal term $\partial z_t/\partial x_t$. Terms that small invite dismissal, and the natural guess is that they cancel among themselves. They do not: they cancel the diagonal term. Adding a constant to every $x_s$, $s \in S$, leaves $z_t$ unchanged, so each row of \Eqref{eq:jacobian} sums to zero exactly. The off-diagonal terms therefore sum to $-\partial z_t/\partial x_t$, individually negligible but collectively as large as the direct path. A sensitivity analysis that perturbs positions one at a time sees only terms of $O(1/|S|)$ and can dismiss the path as negligible; the path is carried by their aggregate.

\item \textbf{Which layers have it is a property of $S$, readable from a definition.} GroupNorm \citep{wu2018group}, InstanceNorm \citep{ulyanov2016instance} and layer normalization \citep{ba2016layer} taken over the sequence put every position of the current input into $S$, as does Conv-TasNet's global layer normalization (gLN). BatchNorm at evaluation replaces $\mu_S$ and $\sigma_S$ with running constants; its Jacobian with respect to the current input is therefore diagonal, $\partial z_t/\partial x_s = (\gamma/\sigma_{\mathrm{run}})\,\delta_{ts}$, and no input-dependent sequence-spanning path survives at inference. Per-token normalization, as transformers \citep{vaswani2017attention} use it, has $S$ inside a single position by construction. Conv-TasNet's cumulative layer normalization (cLN) takes $S = \{s : s \leq t\}$, so the future terms vanish and the past ones do not. It is causal but not local, which is why it appears in Section~\ref{sec:tests} alongside gLN rather than alongside the per-position controls.
\end{enumerate}

\begin{figure}[htbp]
\centering
\includegraphics[width=0.92\linewidth]{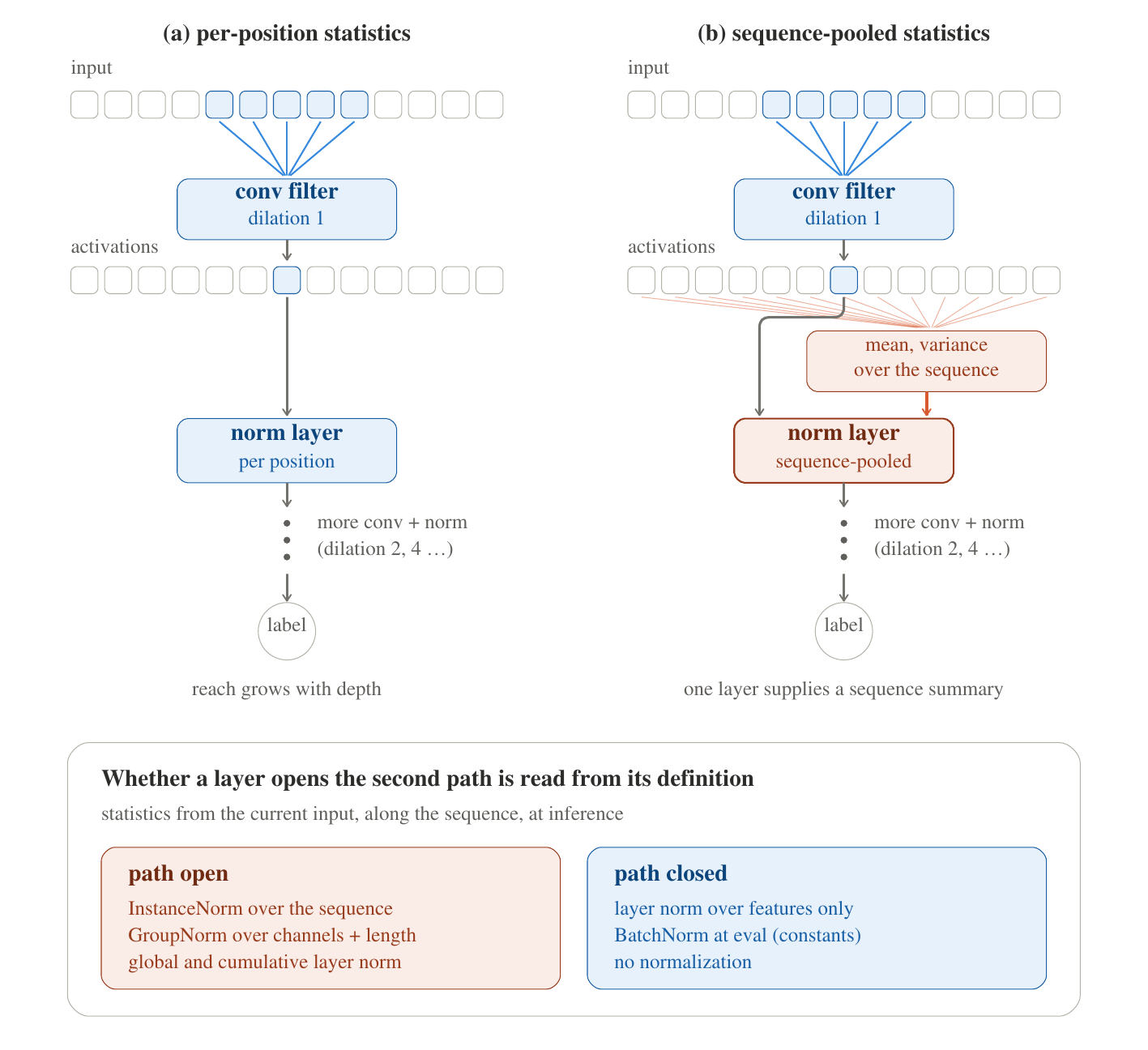}
\caption{Two routes to a sequence labeler's context. The two stacks are identical in architecture, parameter count, data and seed; they differ only in the axes their normalization statistics are pooled over. Each convolution reads a fixed window of its input: the first block's kernel-5 filter is drawn, and reach accumulates over the layers not shown as the dilation doubles. Each normalization layer standardizes its own input activations, and the panels differ in what it standardizes them against: in (a) statistics taken at that position alone, so the layer adds no path between positions; in (b) statistics pooled along the whole activation sequence (faint red lines), so the layer's output at every position depends on every other. The normalization layer in (b) receives two things, the activation being standardized and a summary of all of them, and the second is what no receptive-field calculation accounts for, supplied by a single layer, without depth. The panel below states the exposure criterion and sorts common layers by it. The drawing is schematic: one block is shown, each block being a normalization-and-convolution unit with a residual connection (not drawn), so the normalization layers sit inside the units that ablation removes (Section~\ref{sec:ablation}). The pooled contributions are drawn thin because each term is $O(1/|S|)$, not because their aggregate is small: by \Eqref{eq:jacobian}, they sum to minus the diagonal exactly.}
\label{fig:criterion}
\end{figure}

We call this the exposure criterion: statistics computed from the current input, along the sequence, at inference. A layer that meets it, and a model that contains such a layer, we call exposed. In the rest of the paper, \emph{the path}, unqualified, always means this sequence-spanning path. Figure~\ref{fig:criterion} draws the criterion at the layer where it acts, contrasting the two stacks the experiments of Section~\ref{sec:tests} are built on.

\subsection{What the path can carry} Knowing the path exists says nothing about its worth: that is the second question, and the one that decides whether the path matters in practice. Consider per-position binary labels $y_1,\dots,y_L$ drawn from a stationary symmetric two-state Markov chain with switch probability $p$, observed through per-position evidence of fixed strength $\delta$, the separation between the two classes' mean emissions. Write $n = pL$ for the expected number of label switches in a sequence.\footnote{Exactly $p(L-1)$, since $L$ positions have $L-1$ adjacent transitions; the difference, one part in $L$, is below the approximations everything that uses $n$ carries.} Three quantities decompose the usable context by source, each measured as the best accuracy an optimal predictor reaches from that source alone, under the true prior and emission model (for bounded reach, a conservative bound on it; how each is computed is given in Section~\ref{sec:setup}).

\textbf{Bounded reach} is $R$ positions of context around the target. When runs are long relative to $R$, the $R$ positions usually share a single label, giving an effective separation $\sqrt{R}\,\delta$ and an accuracy $\Phi(\sqrt{R}\,\delta/2)$, with $\Phi$ the standard normal distribution function.

\textbf{Full context} is all $L$ positions, the ceiling no predictor of the observations can pass.

\textbf{The summary alone} is the sequence's class proportion $\bar\pi = L^{-1}\sum_t y_t$, and nothing else. This is what a normalization statistic taken along the sequence encodes to first order; for the Gaussian--Markov process, the pooled mean is exactly an affine function of $\bar\pi$ plus emission noise and nothing else (\ref{app:data}). The pooled statistic is a noisy function of $\bar\pi$ while the oracle receives $\bar\pi$ exactly, so the oracle upper-bounds what this carrier supplies. It is a reference for this process rather than a bound on every global statistic a deeper sequence-pooled network could construct. Where trained networks actually stand relative to it is measured in Section~\ref{sec:tests}, and they sit below it at every density.\footnote{The pooled variance depends on the class mix through $\bar\pi(1-\bar\pi)$, so $\sigma_S$ is a second path to the same sequence-level quantity, carrying $|\bar\pi - \tfrac12|$ without its sign. The bound covers both paths, since each is a function of $\bar\pi$ plus noise.}

The third quantity has a closed form. Since $\operatorname{Var}(y_t)=\tfrac14$ and $\operatorname{corr}(y_t,y_{t+k}) = (1-2p)^k$,

\begin{equation}
\operatorname{Var}(\bar\pi)
 = \frac{1}{4L}\Big[1 + 2\sum_{k=1}^{L-1}\big(1-\tfrac{k}{L}\big)(1-2p)^k\Big]
 \;\longrightarrow\; \frac{1}{4L}\cdot\frac{1-p}{p} \;\approx\; \frac{1}{4n},
\label{eq:var}
\end{equation}
for $n \gg 1$, which lets the sum converge inside the sequence, and $p \ll 1$, which gives the last step. An oracle given $\bar\pi$ predicts the sequence's majority label everywhere,\footnote{This is the per-position Bayes rule given $\bar\pi$: the chain's label-swap symmetry puts the decision threshold at $\bar\pi = \tfrac12$, and its positive correlations ($p < \tfrac12$ throughout) put $P(y_t{=}1 \mid \bar\pi)$ on the same side of $\tfrac12$ as $\bar\pi$.} achieving $\tfrac12 + \mathbb{E}|\bar\pi - \tfrac12|$, and with $\bar\pi$ approximately normal,

\begin{equation}
\mathbb{E}\big|\bar\pi - \tfrac12\big|
 \;\approx\; \sqrt{\tfrac{2}{\pi}\operatorname{Var}(\bar\pi)}
 \;\approx\; \frac{1}{\sqrt{2\pi n}} \;=\; \frac{\derivConst}{\sqrt{n}} .
\label{eq:advantage}
\end{equation}

Nothing in \Eqref{eq:advantage} is fitted. Figure~\ref{fig:decomp} sets it against the exactly computed optima, alongside the other two quantities and the networks that will be measured against them. The approximation holds to within \derivMaxErr\% for $n \geq \derivNMin$; at $n = \derivAtTenN$ the computed value is \derivAtTenObs\ and the prediction \derivAtTenPred\ (Table~\ref{tab:deriv}). Below $n \approx 1$ it must fail: the approximate variance $1/(4n)$ exceeds the largest variance a $[0,1]$-valued quantity can have, while the finite-$L$ sum in \Eqref{eq:var} remains exact and $\bar\pi$ concentrates at the endpoints.

\subsection{Four predictions} If this account is right, then: (P1) where the path is open, reach is worth far less than where it is closed, and the gap is large where labels have long runs; (P2) the length (sequence) axis at inference is what matters: pooling over channels alone changes nothing, pooling over length alone produces the effect, and pooling over length only during training produces none of it; (P3) the gap closes as $n$ grows, since \Eqref{eq:advantage} decays; and (P4) the effect is absent in the two cases where the account predicts none: an architecture that already spans the sequence, and normalization whose statistics are not taken from the current input at inference. Each prediction is tested in Section~\ref{sec:tests}, after Section~\ref{sec:setup} sets out the processes, the models, and the reference quantities.

\section{Measuring what each route is worth}
\label{sec:setup}

Every measurement below is a comparison between models differing in exactly one thing (the receptive field, or which axes the normalization statistics span), trained and evaluated on identical data with identical seeds. Those two are the routes of this section's title: the receptive field, and the path.

\subsection{Generating processes}
\label{sec:processes}

An input sequence is \window\ positions long in all three processes. The receptive field is a property of the model rather than of the data, and varies across our models by a factor of \reachFold. A switch is a position whose label differs from the one before it, and a run is a stretch of positions between two switches. The expected number of switches per sequence, $n$ (Section~\ref{sec:theory}), is the axis the dose--response curves below are swept along, and we call it the switch rate. At the sparse end of the sweep most sequences carry a single label from end to end while at the dense end none do, and the effect we report fades between the two. This answers the charge that processes chosen for computable optima might also be ones in which a sequence's class proportion is unusually informative. The switch rate is swept rather than set, so the design exhibits the regime in which the effect is absent alongside the one in which it is large, and the boundary between them is a prediction rather than a caveat.

Three processes are used rather than one because each answers an objection the others cannot. All three are specified in full in \ref{app:data}. The synthetic process is the one analyzed in Section~\ref{sec:theory}: labels from a symmetric two-state Markov chain with the switch probability set directly, observed through isotropic Gaussian evidence separated by $\delta$. Its optima are exactly computable, so accuracies can be read against an absolute scale rather than only against each other. The simulated genomic process labels each position of a chromosome by which of two populations it derives from: a population-genetic simulator \citep{baumdicker2022efficient} splits two populations and mixes them into mosaics whose segment origins are recorded, so labels are exact. Evidence strength is set by how far the populations have diverged, measured as Hudson's $F_{ST}$ \citep{hudson1992estimation} (zero when their allele frequencies coincide; lower is harder), and the switch rate by the time since mixing. Difficulty is therefore set by a physical parameter with a floor: a model that fails at the hard end has run out of information, not of training. The process ports the demography of a local-ancestry benchmark \citep{tian2026lai}, and its $F_{ST}$ is measured from the simulated panels rather than chosen, matching that benchmark's at all \portLevels\ split times to within \portMaxDiff\ absolute (Table~\ref{tab:port}), so the difficulty axis is fixed by the demography rather than free to be whatever a result requires. The real arm runs the same comparison on \mbox{1000 Genomes} chromosome~22 haplotypes \citep{tgp2015global}, where the sequence statistics are not ours to choose and only the label mosaics are constructed. Training and evaluation sequences come from genomically disjoint segments separated by a buffer, since nearby positions are correlated.

\subsection{Architectures and controls}
\label{sec:arch}

The architecture is a residual dilated stack; keeping the first $k$ blocks gives dilations $1,2,\dots,2^{k-1}$ and receptive fields from \reachMin\ to \reachMax, at \nParams\ parameters when all nine blocks are present (full architectural and training detail in \ref{app:arch}, compute in \ref{app:cost}). The normalization comparison varies only the axes the statistics are pooled over, holding group count, affine parameters and everything else fixed. That the per-position control genuinely closes the path, rather than appearing to, is verified numerically (\ref{app:bounds}); the discipline came from a control of ours that failed, recorded in \ref{app:deff}. Every measurement uses \nSeeds\ seeds and \trainSeq\ training and \evalSeq\ evaluation sequences. Compared models are paired on data and seed.

Truncating the stack removes parameters as well as reach ($k=1$ has \nParamsLo\ parameters against \nParams\ at $k=9$), but capacity cancels from every claim, since each is a comparison between normalizations at the same $k$, where parameter count, depth, optimizer and data are identical and only the pooling axes differ. Capacity also cannot produce the effect: the same \paramFold-fold increase in parameters is worth \normPos\ under per-position normalization and \normGroup\ under sequence-pooled normalization (Table~\ref{tab:axes}; its measure, reach worth, is defined in Section~\ref{sec:measures}). A capacity explanation would have to say why extra parameters help \capacityFold\ more when the normalization statistics happen to exclude the length axis.

\subsection{Reference quantities}
\label{sec:bounds}

The three quantities of Section~\ref{sec:theory} are computed, not estimated, which is what puts measured accuracies on an absolute scale. A fourth item follows them: not a reference quantity but the instrument that reads a receptive field off a model.

\textbf{Full context} is computed by forward--backward over the true chain and the true emission model; equal-covariance Gaussians make the per-position log likelihood ratio a projection, so the recursion is exact.

\textbf{Bounded reach}, $\mathrm{local}(R)$, is a conservative bound rather than the exact centered-window optimum, which would need one forward--backward pass per position. The sequence is instead cut into non-overlapping blocks of $R$ and each position scored from its own block, which is exact at a block center and gives every position at least $R/2$ of context on average, so it should sit below the centered-window optimum, and the distance from a trained network to it therefore overstates what the path is worth at that reach. Networks can nominally exceed it, both because the bound is conservative and because accuracies are seed means. Section~\ref{sec:tests} reports such exceedances at no more than \boundAllAboveMax. Where runs are long relative to $R$, the closed form $\Phi(\sqrt{R}\,\delta/2)$ of Section~\ref{sec:theory} and the forward--backward computation agree to within \closedErr\ for $R$ up to \closedRmax\ (Table~\ref{tab:optima}), an independent check on both.

\textbf{The summary alone} is $\mathbb{E}\max(\bar\pi, 1 - \bar\pi)$ over evaluation sequences: an oracle given the sequence's class proportion predicts its majority label everywhere.

Receptive fields are measured rather than derived wherever the architecture makes the derivation fallible (the U-Net of Section~\ref{sec:tests}): a zero input is passed with gradients enabled, the center output is differentiated, and the receptive field is the extent of the input positions whose gradient is not exactly zero. The comparison is with exact floating-point zero, since absent a path the gradient is identically zero. The input is zero, so the measurement is deterministic, and the nonlinearity's derivative there is nonzero, so no live path is silently zeroed. Run with per-position statistics this returns the convolutional span; run with the same statistics pooled along the sequence it returns the whole sequence, because the normalization is itself a path. The two numbers are the same measurement on the same model under two settings, which is why they can be set against each other.

\subsection{Measurements}
\label{sec:measures}

\textbf{Reach worth} is $\text{accuracy}(k{=}9) - \text{accuracy}(k{=}1)$ for a given normalization, the full stack against the one-block prefix, differenced per seed and then averaged. It is what an architect buys by extending the receptive field, here \reachFold-fold. The \textbf{ratio} is reach worth under per-position statistics divided by reach worth under sequence-pooled statistics.\footnote{The ratio summarizes at the endpoints curves that are reported in full (Figure~\ref{fig:reach}, Table~\ref{tab:grid}), and the endpoints do not manufacture the contrast: on the arms trained at every intermediate prefix, the sequence-pooled family's worth stays within \realPooledEndLo\ to \realPooledEndHi\ at every cut of the real-haplotype sweep while the per-position family's climbs from \realPosEndFirst\ to \realPosEndLast.} A ratio of one says reach is worth as much with the path open as closed; a large ratio says the normalization has already supplied most of what reach would have bought. \textbf{Retention} is the share of above-chance accuracy that survives removing the long-range (reach-enlarging) blocks, $(\text{accuracy}_{\text{removed}} - \tfrac12) / (\text{accuracy}_{\text{full}} - \tfrac12)$; accuracy is bounded below by chance, so a model barely above chance cannot lose much and raw drops are not comparable across difficulty levels. We report retention only where the full model is at least $0.20$ above chance. Where blocks are ablated, they are ablated as a group, all those with dilation at or above a threshold. With residual connections the surviving blocks compensate for any single removal, so per-block importance reads near zero even where the blocks collectively carry the task \citep{morcos2018single, mcgrath2023hydra}. Group removal is also what lets ablation and retraining compare the same architecture, since what remains is the first $k$ blocks, which the retraining sweep already trains from scratch. Throughout, reported uncertainties are standard deviations, across seeds, of the per-seed value of whatever quantity is reported.

\section{Results: testing the predictions}
\label{sec:tests}

Predictions (P1) and (P3) are tested on both generating processes and on real haplotypes, and (P1) additionally in a published architecture; (P2) runs on the genomic process, and (P4)'s two absence cases sit one on each process. Two checks follow the four predictions: that the networks sit at the bounds the decomposition of Section~\ref{sec:theory} sets, and that the account survives an architecture whose pooled extent is not fixed. Per-condition numbers for every result are tabulated in \ref{app:results}.

\subsection{(P1) Reach is worth little when the path is open} Figure~\ref{fig:decomp} places trained networks on the information landscape of the synthetic process. The networks (the dilated stack cut to its first block, receptive field \reachMin\ positions, with statistics pooled along the sequence) track the class-proportion oracle across the whole sweep rather than the bound for their own reach, which stays near chance. The vertical distance between the network and its reach bound is an upper estimate of what the path is worth at that reach.

\begin{figure}[htbp]
\centering
\includegraphics[width=0.98\linewidth]{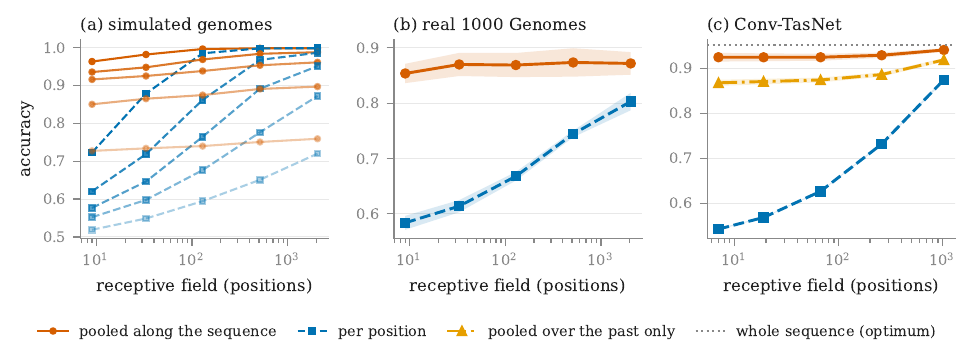}
\caption{(P1) Accuracy against receptive field, with (orange) and without (blue) the sequence-spanning path. Every comparison varies only the axes the normalization statistics are pooled over, holding architecture, parameter count, data and seeds fixed. Reach worth is the accuracy the \reachFold-fold enlargement of the receptive field buys. (a)~the dilated stack on a simulated genomic process at \factLevels\ population divergence levels, $F_{ST}$ from \factFstHi\ (dark, the easiest) to \factFstLo\ (light, the hardest); the sequence-pooled family is nearly flat while the per-position one climbs to meet it, so their reach worth stands in a ratio of \factRatioLo--\factRatioHi\ at all \factLevels\ divergence (difficulty) levels, with no systematic trend across them. (b)~the same stack on real \mbox{1000 Genomes} haplotypes, European and Gujarati Indian panels (\ref{app:data}), where that ratio is \realRatio. (c)~the Conv-TasNet separator, with its recommended gLN (pooled along the sequence), a per-position control, and its causal cLN (pooled over the past only); the gray dotted line is the exact optimum from the whole sequence. cLN and that optimum appear only in (c): cLN is Conv-TasNet's own layer, and only the synthetic process it runs on makes the optimum computable. Points are means over \nSeeds\ seeds, with bands of $\pm$ one standard deviation; (a) omits them for clarity, and in (c) they are narrower than the plotted lines.}
\label{fig:reach}
\end{figure}

Closing the path prices it. On the simulated genomic process, across the \reachFold-fold range of receptive field and a \factFstFold-fold sweep in $F_{ST}$, accuracy moves by \sweepGroupLo\ to \sweepGroupHi\ when normalization statistics span the sequence and by \sweepPosLo\ to \sweepPosHi\ when they do not (Figure~\ref{fig:reach}a, Table~\ref{tab:difficulty}). Taken level by level the ratio is \factRatioLo--\factRatioHi, with no trend across the sweep. Difficulty does not modulate how far normalization substitutes for reach. The families nearly converge at full reach, where a \reachMax-position window already carries nearly all the context the whole sequence does and the path has little left to add. At $F_{ST} = \factFstHi$, the sequence-pooled model reaches \accEasyPooled\ from \reachMin\ positions against the control's \accEasyLocal, and the control needs \accEasyLocalCaught\ positions to match it; at the hardest level, the same pair is \accHardPooled\ against \accHardLocal, the latter barely above chance (Table~\ref{tab:grid}).

\paragraph{On real sequences} Figure~\ref{fig:reach}b takes the comparison to real \mbox{1000 Genomes} CEU/GIH haplotypes ($F_{ST} = \realFst$, \realSwitch\ label switches per sequence): reach across the same range is worth \realGroup\ $\pm$ \realGroupSD\ with sequence-pooled statistics and \realPos\ $\pm$ \realPosSD\ without, a ratio of \realRatio\ (per-seed \realRatioSeedLo--\realRatioSeedHi), even larger than in simulation. The sequence-pooled curve is flat within noise, \realLoAcc\ at receptive field \reachMin\ and \realHiAcc\ at \reachMax: a \reachFold-fold increase in receptive field buys nothing measurable there.

\paragraph{In a published architecture} The Conv-TasNet separator \citep{luo2019convtasnet} is a dilated stack whose receptive field is reported as the extent of its context, and whose recommended non-causal configuration uses global layer normalization (gLN), sequence-spanning at inference. We run that separator unchanged on the synthetic process at \tasSwitch\ switches per sequence, so its exposure can be read against a known optimum. Reach is worth \tasGln\ $\pm$ \tasGlnSD\ under gLN against \tasPos\ $\pm$ \tasPosSD\ under the per-position control, a ratio of \tasGlnRatio\ (per-seed \tasRatioSeedLo--\tasRatioSeedHi; Figure~\ref{fig:reach}c); the model reaches \tasGlnLo\ at a receptive field of only \tasReachLo\ positions, against an exact optimum of \tasCeiling\ from all \window. Its causal variant, cumulative layer normalization, pools only over the past yet shows a ratio of \tasClnRatio. With runs this long, the past of a sequence already carries most of what the whole sequence says about its class proportion, so a one-sided running summary substitutes nearly as well as a whole-sequence one. Streaming deployments are therefore not exempt. Causality is preserved, locality is not, and the nominal receptive field no longer bounds the context available to the model. What the path is worth on speech separation itself is a separate study (Section~\ref{sec:discussion}).

\subsection{(P2) The length axis, at inference, is what produces the effect} Table~\ref{tab:axes} gives reach worth for all five normalizations. Pooling over channels alone is indistinguishable from no normalization (\normPos\ against \normNone); pooling over length alone produces the whole effect (\normInst), matching GroupNorm (\normGroup). BatchNorm, which pools over length in training but uses running statistics at evaluation, behaves as though it had no path (\normBatch), consistent with the observation of \citet{pfrommer2025spooky} and with the exposure criterion. Constants carry nothing from the current input, whatever axes they were once pooled over.

\begin{table}[t]
\centering
\caption{(P2) Reach worth (the accuracy gained by enlarging the receptive field) by the axes the normalization statistics span. The third column is the exposure criterion of Section~\ref{sec:theory} (a path exists exactly when the statistics are computed from the current input, along the sequence, at inference) applied to each layer's definition, before any of these numbers were measured. BatchNorm is the row where training and inference differ, pooling along the sequence while training but substituting running constants at evaluation. Thus, the criterion predicts no path, and it behaves like the rows without one (none and per position). All rows are at one divergence level, $F_{ST} = \normFst$. Means over \nSeeds\ seeds.}
\label{tab:axes}
\resizebox{\ifdim\width>\linewidth\linewidth\else\width\fi}{!}{%
\begin{tabular}{llcr}
\toprule
normalization & statistics pooled over & path at inference & reach worth \\
\midrule
none            & ---                               & no  & \normNone\ $\pm$ \normNoneSD \\
per position    & channels                          & no  & \normPos\ $\pm$ \normPosSD \\
instance        & length                            & yes & \normInst\ $\pm$ \normInstSD \\
group           & channels and length               & yes & \normGroup\ $\pm$ \normGroupSD \\
batch           & batch and length; running at eval & no  & \normBatch\ $\pm$ \normBatchSD \\
\bottomrule
\end{tabular}}
\end{table}

\subsection{(P3) The gap closes as labels switch more often} 
\begin{figure}[htbp]
\centering
\includegraphics[width=0.98\linewidth]{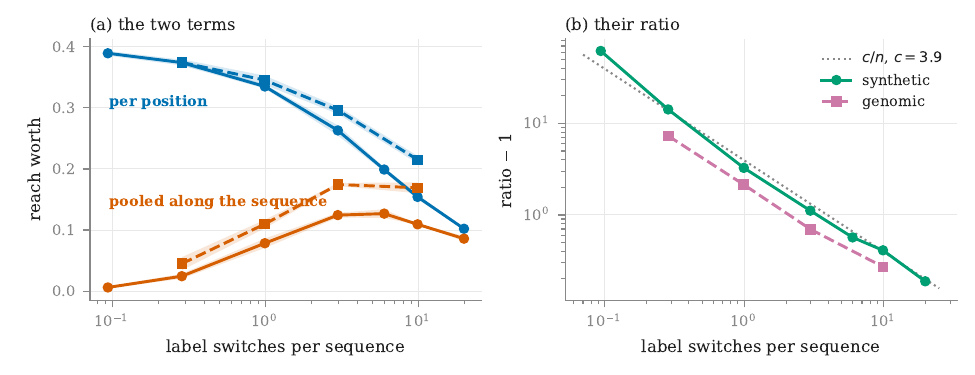}
\caption{(P3) Normalization substitutes for reach (the receptive field) less and less as labels switch more often. Reach worth is the accuracy gained by enlarging the receptive field, and the ratio is reach worth without the sequence-spanning path divided by reach worth with it, so a large ratio means the normalization had already supplied what reach would have bought. (a)~the two quantities the ratio is built from, reach worth with the path (orange) and without it (blue). Both depend on the switch rate, and over most of the panel in opposite directions: as the summary degrades (more switches) the path supplies less of the context, so reach is worth more to a sequence-pooled model; a per-position model has no path at all, so its context is its receptive field and more switches leave less within it to integrate. (b)~their ratio minus one, both axes logarithmic; the dotted reference is $c/n$, which is a slope of $-1$ on these axes. Solid lines are the synthetic process, dashed the genomic one, and the bands in (a) are $\pm$ one standard deviation over \nSeeds\ seeds.}
\label{fig:dose}
\end{figure}

Raising the switch rate with everything else fixed, the ratio of reach worth with the path closed to reach worth with it open falls monotonically on both processes (Figure~\ref{fig:dose}b). On the genomic process it is \genDoseRatio\ at \genDoseN\ switches per sequence respectively. On the synthetic process it runs from \synRatioMax\ to \synRatioMin\ across a \synDoseFold-fold range of switch rate, from \synNMin\ to \synNMax. The two processes can be compared point for point, since the synthetic one was also run at the switch densities the genomic one happens to produce: at \matchedSynN\ switches per sequence it gives \matchedSynRatios, against the genomic process's \matchedGenRatios\ at the same densities. The same shape comes from two generating processes that share no mechanism, converging where labels switch often. At the highest switch rates the two routes are worth about the same and a receptive-field justification is approximately right: the effect is a property of long label runs, not of sequence labeling in general.

Empirically, the ratio follows $(\text{ratio}-1) \approx c/n$, with $c = \lawC$ and a relative spread of \lawSpread\%\ across the full range while $(\text{ratio}-1)$ itself varies \synSpanRatio-fold (Table~\ref{tab:dose}). Fitting the exponent freely gives $\lawAlphaSyn \pm \lawAlphaSynSE$ on the synthetic process and $\lawAlphaGen \pm \lawAlphaGenSE$ on the genomic one, both consistent with $1$; imposing $1/\sqrt{n}$ instead leaves a constant that varies by a factor of \lawRootFold\ across the sweep, so the data separate the two forms. The exponent needs stating carefully. \Eqref{eq:advantage} gives $1/\sqrt{n}$ for the path's value, but the ratio is not the path's value. Its numerator, reach worth without the path, carries a dependence on $n$ of its own, because shorter runs leave less coherent signal within any receptive field (Figure~\ref{fig:dose}a, where the two terms of the ratio move in opposite directions over most of the panel). The bounds fix where the regimes lie and that the ratio falls toward one, but not the exponent: substituting the bounds and optima for the trained accuracies gives an idealized version of both reach-worth curves, and the product $(\text{ratio}-1)\,n$ formed from their ratio varies \idealCFold-fold where the measured one varies \lawSpread\%. We therefore report \Eqref{eq:advantage} as derived and the $1/n$ as an empirical regularity, whose exponent transfers across processes and whose constant does not.

\subsection{(P4) The substitution is absent where the account says it must be} The account has two halves, a criterion for which layers open a path and a closed form for what an open one carries. Each yields a falsifiable prediction of absence, and both hold. A transformer, whose attention spans the input at every depth, differs by only \xfDiffLo--\xfDiffHi\ between sequence-pooled and per-position normalization, while sitting within \xfGap\ of the exact optimum, at each of \xfDensities\ switch densities (Table~\ref{tab:transformer}): with the context already available, the path has nothing left to supply. The same runs certify that the task is solvable to near the optimum by a model that can see the whole sequence, so when a short-reach stack falls short of its own bound, the shortfall belongs to what it can reach rather than to task difficulty or training. The second case is BatchNorm, already reported under (P2): it pools along the sequence while training, yet shows no substitution, because running statistics sever the path exactly where it would be used. An account that merely correlated normalization with reduced dependence on reach would predict an effect wherever normalization appears; this one predicts precisely where it will not, and is right in both cases for different reasons, one architectural, one an implementation detail of the layer.

\subsection{Networks operate at the limits their information allows} One further check ties the predictions back to Section~\ref{sec:theory}: the networks are not merely ordered as predicted but sit at the limits the decomposition sets. Per-position models attain \boundLocalLo--\boundLocalHi\% of the local bound for their receptive field at the shortest reach, as a share of above-chance accuracy, at all \boundNDensities\ switch densities. Across all \boundAllCells\ combinations of receptive field and density, their mean absolute gap to the bound is \boundAllMean. In \boundAllAbove\ of those cells the network is nominally above it, by at most \boundAllAboveMax, as a conservative bound and finite-sample means lead one to expect (Table~\ref{tab:resid}). At \reachMin\ positions the bound is tighter still, and every cell of that column sits below it, by at most \boundLocalRawMax. Sequence-pooled models are where no bounded-reach predictor could be: at the sparsest density, a model reaching \reachMin\ positions exceeds the local bound for that reach by \boundGroupExcess\ in accuracy, \boundExcessFold\ times the largest slack anywhere in the grid. Across the sweep they track the proportion oracle instead, attaining \boundGlobalHi\% of it where labels rarely switch and \boundGlobalLo\% at \synNMax\ switches per sequence (Figure~\ref{fig:decomp}). Some shortfall is required: the oracle receives $\bar\pi$ exactly, while the pooled statistic carries it over a noise floor, and \ref{app:data} measures that carrier degrading over the same sweep. The comparison in (P1) is therefore not between a good model and a bad one. A bounded-reach network with per-position statistics is at the information limit of what its inputs allow, and the ratio compares a model at its limit against one with an extra path.

\subsection{The account survives where the pooled extent shrinks with depth} A U-Net \citep{ronneberger2015unet} halves the sequence at each depth, so a statistic at depth $d$ pools over $L/2^d$ values rather than $L$. The criterion says a path exists at every depth, while the $O(1/|S|)$ magnitude argument says the values it is built from grow scarce. This is where the two halves of the account could come apart, and it is the architecture \citet{pfrommer2025spooky} name as the practical risk. They do not come apart. Measured by autograd (Section~\ref{sec:bounds}), the receptive field with per-position statistics is \unetRFLo\ to \unetRFHi\ positions across depth, a \unetRFFold-fold range. With the same statistics pooled along the sequence, the measurement returns \unetWindow\ at every depth, the whole sequence. That is the claim of this paper obtained as an observation rather than an argument. Sequence-pooled accuracy is flat to within \unetFlat\ while the pooled count falls \unetPoolFold-fold, because even the deepest statistic still pools \unetValuesDeep\ values, far from the count at which its noise floor would swamp what it carries. Depth moves reach and pooled extent together, which is why this architecture cannot separate them, but the confound runs against the finding: over this range reach grows \unetRFFold-fold while the pooled count falls, and both changes should reduce the path's advantage over convolutional reach rather than preserve it. What the path is worth follows the dilated stacks. At \unetSwSparse\ switches per sequence the U-Net's reach worth stands in a ratio of \unetRatioSparse, beside \synRatioMatched\ for the dilated stack at the same density. The shallowest sequence-pooled U-Net attains \unetPctOracle\%\ of the proportion oracle from \unetRFLo\ positions of convolutional reach, while its per-position control sits between the bounds for \reachMin\ and \reachBracketHi\ positions, which bracket that reach. Raising the switch rate to \unetSwDense\ collapses the ratio to \unetRatioDense\ (Table~\ref{tab:unet}), the decay (P3) predicts.

\section{The consequence for attribution}
\label{sec:ablation}

Two estimates can answer what a component contributes. One deletes it from a trained network and re-evaluates; the other builds the same architecture without it and trains from the start. Both are called ablation in practice; below, \emph{ablation} exclusively means the first and \emph{retraining} the second. Where the removed component both extends the receptive field and carries normalization whose statistics span the sequence, ablation does not just shorten the reach. It closes part of the path at the same time, and the network it leaves behind, its weights fit to the old routing, cannot obtain the summary through what remains. A network trained without the component does obtain it, from the normalization layers it does have. The accuracy ablation finds missing is therefore attributed to the component, although most of that context was reachable without it. This section measures how large that error is, and separates it from the ordinary disagreement the two estimates show even with no path to sever: a retrained network adapts to the removal, and an ablated network's surviving components compensate for it.

\begin{figure}[htbp]
\centering
\includegraphics[width=0.98\linewidth]{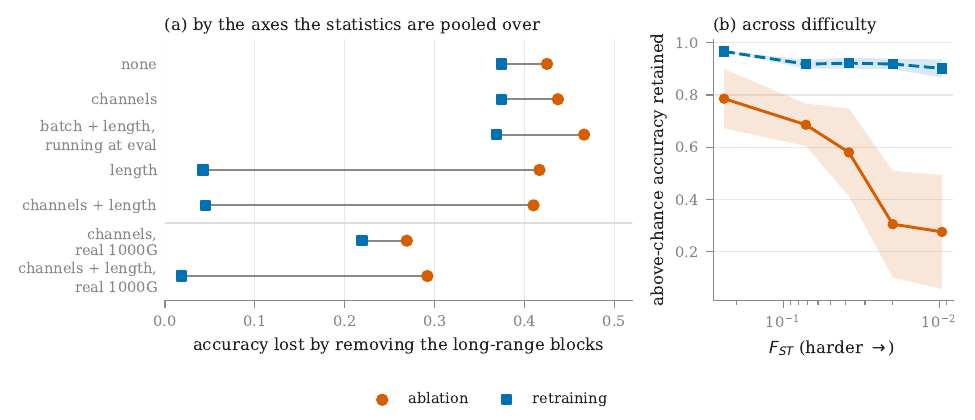}
\caption{Two ways of estimating the contribution of a dilated convolutional stack's long-range blocks (blocks that enlarge the receptive field; Section~\ref{sec:setup}): ablating them from a trained network and re-evaluating it with no fine-tuning (orange) against training the same architecture without them from the start (blue). (a)~Accuracy lost by removing the blocks, so further right is a larger reported loss, and ablation sits to the right of retraining in every row. The gap between them depends only on which axes the normalization statistics are pooled over: the five rows above the rule share a task, an architecture, seeds and data, and differ only in the normalization layer, while the two below repeat the contrast on real haplotypes. Where the statistics do not span the sequence at inference (including BatchNorm, which spans it only while training), the two measurements nearly agree (there, the blocks' reach was genuinely needed, and retraining without sequence-spanning statistics does not recover much). Where they do span it, the two differ by \disOpenLo--\disOpenHi\ times: the difference is context that sequence-spanning statistics can recover and reach alone cannot. (b)~The sequence-pooled (GroupNorm) configuration, at a shallower cut that removes only the blocks with dilation $\geq 8$: the two estimates disagree twice over, about how much the blocks matter and about whether that changes with task difficulty. Retention, the share of above-chance accuracy that survives removal, is far lower by ablation than by retraining at every difficulty, and it falls steeply as the task hardens where retraining finds it relatively flat. Points in (a) are means over \nSeeds\ seeds; the bands in (b) are $\pm$ one standard deviation.}
\label{fig:ablation}
\end{figure}

Because each block enters as $h \leftarrow h + f(h)$, removing it is exactly the identity map, so block ablation here needs no surgery, no fine-tuning and no replacement value. It is nonetheless biased by a large factor. On the simulated genomic process, pooled over its \factLevels\ divergence levels and with normalization pooled along the sequence, ablating the reach-enlarging blocks from the trained network costs \ablCostOpen\ of accuracy where training a network that never had them costs \retCostOpen: a ratio of \ablRatioOpen\ (per level in Table~\ref{tab:ablcost}). On real haplotypes, ablation reports \realAbl\ $\pm$ \realAblSD\ against \realRet\ $\pm$ \realRetSD\ by retraining, a ratio of \realAblRatio\ (the bottom rows of Figure~\ref{fig:ablation}a). Ablation is also noisier: its seed-to-seed spread is \realNoiseFold\ times as large. It is therefore both biased and imprecise, and the bias is the larger problem: read as what the blocks are worth, its number bills them for the network's sequence-spanning path.

The size of the ablation estimate is not the only thing that goes wrong; so is its trend across task difficulty. Analyses of this kind usually report a single cut that removes the long-range tail of the model and ask how much of its skill survives. To mirror that practice, ours cuts the blocks with dilation $\geq 8$, a removal shallower than the one above. At that cut, ablation puts retention at \cutAblRet\ $\pm$ \cutAblSD: about half of what the model can do above chance appeared to depend on the removed blocks. Retraining puts it at \cutRetRet\ $\pm$ \cutRetSD: almost none of it actually depended on them. The two estimates also tell different stories about difficulty (Figure~\ref{fig:ablation}b). Read by ablation, reliance on the long-range blocks appears to grow sharply as the task gets harder, retention falling from \cutAblHi\ at $F_{ST} = \factFstHi$ to \cutAblLo\ at \factFstLo. Read by retraining, it barely moves: \cutRetHi\ to \cutRetLo\ over the same range. An analysis built on the first curve would report that harder tasks lean harder on the long-range architecture, which by the retrained estimate they do not, and would report it with error bars \cutNoiseFold\ times wider than necessary.

We attribute the disagreement to what ablation removes besides the blocks: each block's normalization layers go with its convolutions, and with them part of the sequence-spanning path, so the path's contribution lands on the blocks' bill. This explanation has to be separated from a familiar alternative, compensation among the surviving components \citep{mcgrath2023hydra, hooker2019benchmark}, and one controlled comparison separates them (Table~\ref{tab:dissect}, Figure~\ref{fig:ablation}a): holding the task, the architecture, the seeds and the data fixed, and changing only which axes the normalization statistics span. Where no statistic spans the sequence at inference, there is no path to remove, and the two estimates differ by \disClosedLo--\disClosedHi\ times. That band is what adaptation and compensation together produce. Both costs there are large as well as close, since with nothing to substitute for it the blocks' reach is genuinely needed. Where a statistic does span it, they differ by \disOpenLo--\disOpenHi\ times. The factor of \disBandFold\ between the two bands is attributable to the path, because nothing else differs between the rows. Nor is the contrast a property of a single difficulty level: pooled across the whole sweep, the ablated cost is only \ablRatioClosed\ times the retrained one with per-position statistics, against the \ablRatioOpen\ times reported above with the length axis restored (Table~\ref{tab:ablcost}).

\begin{table}[t]
\centering
\caption{Removing the long-range blocks of a dilated convolutional stack (every block that enlarges the receptive field; Section~\ref{sec:setup}) from the trained network costs accuracy: column 3, the cost with no retraining. Column 4 retrains the same reduced architecture from scratch, and how much of the drop it recovers depends on the one thing that varies across the rows, whether the normalization statistics give an inference-time path along the sequence. Without one (top three rows), retraining recovers little and the two costs nearly agree, differing by \disClosedLo--\disClosedHi\ times (the gap adaptation and compensation alone produce): there, the blocks' reach was genuinely needed. With one (bottom two rows), the costs differ by \disOpenLo--\disOpenHi\ times: retraining recovers nearly all of the drop, because the network re-obtains the sequence summary through the normalization layers it keeps. The inference-time path along the sequence, then, can supply most of the context the blocks were credited with, without any of their reach; the excess of the ablated cost over the retrained one, beyond the adaptation-and-compensation gap the top rows calibrate, measures the sequence-spanning path severed with the removed blocks (each block carries its own normalization layers), not the blocks themselves. The ``by retraining'' cost is the full network's accuracy minus that of a network trained from scratch without the long-range blocks, which is the same difference Table~\ref{tab:axes} reports as reach worth (the accuracy gained by enlarging the receptive field). All rows are at one divergence level, $F_{ST} = \normFst$. The ratios pooled across the difficulty sweep are in Table~\ref{tab:ablcost}. Means over \nSeeds\ seeds.}
\label{tab:dissect}
% Generated by paper/make_numbers.py. Do not edit.
\begin{tabular}{llrrr}
\toprule
normalization & statistics pooled over & by ablation & by retraining & ratio \\
\midrule
none & --- & $+0.4249$ & $+0.3740$ & $1.1$ \\
per position & channels & $+0.4371$ & $+0.3741$ & $1.2$ \\
batch & batch and length, running at eval & $+0.4662$ & $+0.3687$ & $1.3$ \\
instance & length & $+0.4165$ & $+0.0424$ & $9.8$ \\
group & channels and length & $+0.4099$ & $+0.0452$ & $9.1$ \\
\bottomrule
\end{tabular}

\end{table}

\paragraph{What to do instead of trusting the ablation number} The check we recommend is the one used throughout this paper, and it costs one additional training run: repeat the attribution with the length axis removed from the normalization statistics, closing the sequence-spanning path. If the conclusion moves, the original number was also measuring the normalization path rather than the component alone. Where the removed component carries sequence-spanning normalization and retraining without it is affordable, retraining is the better measurement, because it does not sever the path along with the component. On our task it is also the steadier one. That is not a general ranking of ablation and retraining: the two answer different questions, and ablation remains the right one to ask where removing a component removes nothing else.

\section{Discussion}
\label{sec:discussion}

\subsection{When the effect matters} The check that closes Section~\ref{sec:ablation} generalizes past attribution (retrain once with the length axis removed from the normalization statistics; the model is its own control, and how far any conclusion moves is how much of its context came from normalization). Whether it is worth running can be settled in advance. \Eqref{eq:advantage} depends on $n$, the number of switches an input sequence contains, rather than on sequence length, and $n$ is estimable as sequence length divided by typical label run. Below one switch per sequence a model sits deep inside the regime, near three the substitution is still worth about a factor of two, and by ten it has all but gone (Table~\ref{tab:dose}). The empirical $c/n$ law sharpens this to a rate, halving the switch rate roughly doubling the ratio's excess over one, and the factor itself where the substitution is large, though its constant is not transferable across settings (Section~\ref{sec:tests}).

\subsection{Which architectures are exposed} That the exposure criterion is narrow is what makes it useful: it names a property of a layer, so a reader settles the question by reading a definition rather than running an experiment. BatchNorm escapes it by an implementation detail (running statistics replace minibatch ones at evaluation). The escape is contingent on that detail. A deployment that leaves BatchNorm in training mode at inference, or a test-time adaptation method that recomputes statistics on the data being predicted, meets the criterion.

Checked against released source rather than papers, the criterion partitions \surveyN\ published models in \surveyDomains\ domains whose labels come in long runs: \surveyDomainList\ (Table~\ref{tab:survey}; the released survey file records each reading, pinned to the commit read, with exclusions and reasons). \surveyExposed\ are exposed: ASFormer \citep{yi2021asformer}, whose attention modules apply InstanceNorm1d over the input at inference, and pyannote.audio's SincNet frontend \citep{bredin2020pyannote}, where what the normalization bypasses is the frontend's receptive field rather than the model's context, since a recurrent layer follows. The other \surveyUnexposed\ are not, for the two reasons the criterion names: batch statistics frozen at evaluation, or layer normalization over the feature axis at each position. Enformer \citep{avsec2021effective} and Basenji \citep{kelley2018sequential}, both motivated by the genomic distance they integrate, are both unexposed at inference, against our expectation. Exposure follows neither the domain nor the reported reach (both exposed models sit in domains that also contain unexposed ones), so the property must be read from the normalization layer itself, and read from code rather than papers. One diarization model specifies an input layer normalization its released code instantiates and never applies. Had we cited the equation we would have called it exposed, and it is not. The partition even reappears inside a single model: Mamba \citep{gu2024mamba} normalizes per position and does not meet the criterion, while S4 \citep{gu2022s4} is unexposed under its default layer normalization yet exposed under its instance and group options, so one configuration string decides whether the path exists. These are readings of source, not measurements: a met criterion is necessary for the substitution but says nothing of its size, and settling the size for any of these models means running it on a task whose optima are known, as done here for Conv-TasNet.

\subsection{An uncomfortable corollary} Exposure is not a property anyone chose. A practitioner who leaves BatchNorm, as many do for small or variable batches, for variable-length inputs, or for the documented pathologies of batch statistics \citep{wu2018group, brock2021high}, usually lands on GroupNorm or InstanceNorm taken along the sequence, and acquires the path as a side effect of the move. The discomfort is that although the move is good practice, for reasons that have nothing to do with context, a receptive-field justification can stop being true in a change that looks like housekeeping.

\subsection{Scope and limitations} What this paper claims is bounded on both sides. Not that receptive field is unimportant: where the path is closed, reach carries the task almost entirely (Figure~\ref{fig:reach}), and where labels switch often the two routes are worth about the same. Not that models using the path are worse; where labels run long they are better, having found information their architecture nominally denies them. The claim is that the reasons given for these architectures, and the measurements offered in support, credit the receptive field with context that the normalization supplies, and that credit is the premise behind dilation schedules, behind the locality guarantees of streaming systems, and behind the ablations that decide what survives into the next design. In the regime we measure it is off by roughly an order of magnitude.

The claim is conditional and the condition is measured rather than assumed: the sweep of Figure~\ref{fig:dose} crosses the boundary, and everything here is scoped to sequence labeling with long label runs. We have not tested label processes that are structured rather than Markov, and two violations can be named in advance. A task whose sequences are class-balanced by construction gives the summary no variance to carry. A statistic pooled over the whole sequence is also blind to arrangement, so a task that depends on which segment comes first should gain little, with cLN's cumulative statistic the exception, since what it has pooled by position $t$ is what came before. Nor have we tested inputs whose length at inference differs from training. The derived half is narrower still. \Eqref{eq:advantage} is a statement about binary labeling, where the summary is the scalar $\bar\pi$. With $K$ classes the summary is a $(K-1)$-dimensional proportion vector and the closed form would need re-deriving, and for regression there is no class proportion at all. The exposure criterion is untouched by this, being a claim about a layer rather than a label alphabet; what does not generalize is the closed form, the part that lets accuracies be read on an absolute scale. The Conv-TasNet results use that architecture on the controlled process, and what share of its speech-separation performance comes from gLN would require training on a separation corpus against separation metrics, a study in its own right. The real-data arm uses one chromosome and one population pair, its three replicates sharing whatever is particular to that chromosome. All measurements use \nSeeds\ seeds, few enough that the ablation-based retention is read as an order of magnitude rather than a precise value, and the largest ratios, whose sequence-pooled denominators are small, are reported with their per-seed ranges. Hyperparameters are held to one configuration, though a control at its computed bound leaves tuning nothing to improve. The constant $c$ in the switch-rate regularity varies across settings in a way we do not explain.
% Limitations commented out 2026-08-19 as duplicates, recoverable here:
% - local(R) is a blocked bound, used only as a bound (stated in Section 4.3).
% - The survey reports the criterion, not effect size (stated at the end of 7.2).
% - A nonzero Jacobian is existence, not amount (Section 3.2's opening distinction).

\section{Conclusion}

For a convolutional sequence labeler whose normalization statistics are pooled along the sequence, the receptive field computed from the convolutions is not a good description of the context the model uses, and where labels come in long runs, it is not even the dominant contributor. Two routine practices inherit the error: justifying an architecture by the span it reaches, and attributing behavior to a model's reach-enlarging components by removing them from a trained network. In the regime we measure, both overstate what the receptive field contributes by roughly an order of magnitude, and the overstatement disappears once the normalization statistics are taken per position, which is also the cheapest diagnostic available, at one additional training run. What a practitioner can settle in advance is not the size of the effect, which the bounds do not predict, but whether they are in the regime at all. That takes two checks and no experiment: whether the normalization statistics are computed from the current input along the sequence at inference, which follows from the layer's definition, and whether the labels come in long runs, which follows from the task.

\section*{Data availability}

Everything needed to regenerate the results is released at \url{https://github.com/qtianreal/sequence-pooled-normalization}: the two simulated generating processes (Gaussian--Markov and coalescent ancestry), the architectures, the bounds computation, the analysis scripts, and the result files for all \numExperiments\ training runs. From these, each table, each figure, and every number in the prose can be regenerated and checked against the paper. The real-data arm additionally requires the public \mbox{1000 Genomes} chromosome~22 release, which we do not redistribute. No inference about individuals is made or attempted, and no new data was collected.

\section*{Acknowledgment}

This work was supported in part by the National Science Foundation (NSF) under Award No. 2412285.

\bibliographystyle{elsarticle-harv}
\bibliography{paper}

\begin{thebibliography}{46}
\expandafter\ifx\csname natexlab\endcsname\relax\def\natexlab#1{#1}\fi
\providecommand{\url}[1]{\texttt{#1}}
\providecommand{\href}[2]{#2}
\providecommand{\path}[1]{#1}
\providecommand{\DOIprefix}{doi:}
\providecommand{\ArXivprefix}{arXiv:}
\providecommand{\URLprefix}{URL: }
\providecommand{\Pubmedprefix}{pmid:}
\providecommand{\doi}[1]{\href{http://dx.doi.org/#1}{\path{#1}}}
\providecommand{\Pubmed}[1]{\href{pmid:#1}{\path{#1}}}
\providecommand{\bibinfo}[2]{#2}
\ifx\xfnm\relax \def\xfnm[#1]{\unskip,\space#1}\fi
%Type = Article
\bibitem[{Avsec et~al.(2021)Avsec, Agarwal, Visentin, Ledsam, Grabska-Barwinska, Taylor, Assael, Jumper, Kohli and Kelley}]{avsec2021effective}
\bibinfo{author}{Avsec, {\v{Z}}.}, \bibinfo{author}{Agarwal, V.}, \bibinfo{author}{Visentin, D.}, \bibinfo{author}{Ledsam, J.R.}, \bibinfo{author}{Grabska-Barwinska, A.}, \bibinfo{author}{Taylor, K.R.}, \bibinfo{author}{Assael, Y.}, \bibinfo{author}{Jumper, J.}, \bibinfo{author}{Kohli, P.}, \bibinfo{author}{Kelley, D.R.}, \bibinfo{year}{2021}.
\newblock \bibinfo{title}{Effective gene expression prediction from sequence by integrating long-range interactions}.
\newblock \bibinfo{journal}{Nature Methods} \bibinfo{volume}{18}, \bibinfo{pages}{1196--1203}.
%Type = Article
\bibitem[{Ba et~al.(2016)Ba, Kiros and Hinton}]{ba2016layer}
\bibinfo{author}{Ba, J.L.}, \bibinfo{author}{Kiros, J.R.}, \bibinfo{author}{Hinton, G.E.}, \bibinfo{year}{2016}.
\newblock \bibinfo{title}{Layer normalization}.
\newblock \bibinfo{journal}{arXiv} \DOIprefix\doi{10.48550/arXiv.1607.06450}. \bibinfo{note}{preprint}.
%Type = Article
\bibitem[{Bai et~al.(2018)Bai, Kolter and Koltun}]{bai2018tcn}
\bibinfo{author}{Bai, S.}, \bibinfo{author}{Kolter, J.Z.}, \bibinfo{author}{Koltun, V.}, \bibinfo{year}{2018}.
\newblock \bibinfo{title}{An empirical evaluation of generic convolutional and recurrent networks for sequence modeling}.
\newblock \bibinfo{journal}{arXiv} \DOIprefix\doi{10.48550/arXiv.1803.01271}. \bibinfo{note}{preprint}.
%Type = Article
\bibitem[{Baumdicker et~al.(2022)Baumdicker, Bisschop, Goldstein, Gower, Ragsdale, Tsambos, Zhu, Eldon, Ellerman, Galloway, Gladstein, Gorjanc, Guo, Jeffery, Kretzschumar, Lohse, Matschiner, Nelson, Pope, Quinto-Cortes, Rodrigues, Saunack, Sellinger, Thornton, van Kemenade, Wohns, Wong, Gravel, Kern, Koskela, Ralph and Kelleher}]{baumdicker2022efficient}
\bibinfo{author}{Baumdicker, F.}, \bibinfo{author}{Bisschop, G.}, \bibinfo{author}{Goldstein, D.}, \bibinfo{author}{Gower, G.}, \bibinfo{author}{Ragsdale, A.P.}, \bibinfo{author}{Tsambos, G.}, \bibinfo{author}{Zhu, S.}, \bibinfo{author}{Eldon, B.}, \bibinfo{author}{Ellerman, E.C.}, \bibinfo{author}{Galloway, J.G.}, \bibinfo{author}{Gladstein, A.L.}, \bibinfo{author}{Gorjanc, G.}, \bibinfo{author}{Guo, B.}, \bibinfo{author}{Jeffery, B.}, \bibinfo{author}{Kretzschumar, W.W.}, \bibinfo{author}{Lohse, K.}, \bibinfo{author}{Matschiner, M.}, \bibinfo{author}{Nelson, D.}, \bibinfo{author}{Pope, N.S.}, \bibinfo{author}{Quinto-Cortes, C.D.}, \bibinfo{author}{Rodrigues, M.F.}, \bibinfo{author}{Saunack, K.}, \bibinfo{author}{Sellinger, T.}, \bibinfo{author}{Thornton, K.}, \bibinfo{author}{van Kemenade, H.}, \bibinfo{author}{Wohns, A.W.}, \bibinfo{author}{Wong, Y.}, \bibinfo{author}{Gravel, S.}, \bibinfo{author}{Kern, A.D.}, \bibinfo{author}{Koskela, J.}, \bibinfo{author}{Ralph, P.L.}, \bibinfo{author}{Kelleher, J.}, \bibinfo{year}{2022}.
\newblock \bibinfo{title}{Efficient ancestry and mutation simulation with msprime 1.0}.
\newblock \bibinfo{journal}{Genetics} \bibinfo{volume}{220}.
%Type = Inproceedings
\bibitem[{Bredin et~al.(2020)Bredin, Yin, Coria, Gelly, Korshunov, Lavechin, Fustes, Titeux, Bouaziz and Gill}]{bredin2020pyannote}
\bibinfo{author}{Bredin, H.}, \bibinfo{author}{Yin, R.}, \bibinfo{author}{Coria, J.M.}, \bibinfo{author}{Gelly, G.}, \bibinfo{author}{Korshunov, P.}, \bibinfo{author}{Lavechin, M.}, \bibinfo{author}{Fustes, D.}, \bibinfo{author}{Titeux, H.}, \bibinfo{author}{Bouaziz, W.}, \bibinfo{author}{Gill, M.P.}, \bibinfo{year}{2020}.
\newblock \bibinfo{title}{pyannote.audio: Neural building blocks for speaker diarization}, in: \bibinfo{booktitle}{IEEE International Conference on Acoustics, Speech and Signal Processing}.
%Type = Inproceedings
\bibitem[{Brock et~al.(2021)Brock, De, Smith and Simonyan}]{brock2021high}
\bibinfo{author}{Brock, A.}, \bibinfo{author}{De, S.}, \bibinfo{author}{Smith, S.L.}, \bibinfo{author}{Simonyan, K.}, \bibinfo{year}{2021}.
\newblock \bibinfo{title}{High-performance large-scale image recognition without normalization}, in: \bibinfo{booktitle}{International Conference on Machine Learning}.
%Type = Inproceedings
\bibitem[{Desplanques et~al.(2020)Desplanques, Thienpondt and Demuynck}]{desplanques2020ecapa}
\bibinfo{author}{Desplanques, B.}, \bibinfo{author}{Thienpondt, J.}, \bibinfo{author}{Demuynck, K.}, \bibinfo{year}{2020}.
\newblock \bibinfo{title}{{ECAPA-TDNN}: Emphasized channel attention, propagation and aggregation in {TDNN} based speaker verification}, in: \bibinfo{booktitle}{Interspeech}, pp. \bibinfo{pages}{3830--3834}.
%Type = Inproceedings
\bibitem[{Farha and Gall(2019)}]{farha2019mstcn}
\bibinfo{author}{Farha, Y.A.}, \bibinfo{author}{Gall, J.}, \bibinfo{year}{2019}.
\newblock \bibinfo{title}{{MS-TCN}: Multi-stage temporal convolutional network for action segmentation}, in: \bibinfo{booktitle}{IEEE Conference on Computer Vision and Pattern Recognition}.
%Type = Inproceedings
\bibitem[{Fujita et~al.(2019)Fujita, Kanda, Horiguchi, Xue, Nagamatsu and Watanabe}]{fujita2019eend}
\bibinfo{author}{Fujita, Y.}, \bibinfo{author}{Kanda, N.}, \bibinfo{author}{Horiguchi, S.}, \bibinfo{author}{Xue, Y.}, \bibinfo{author}{Nagamatsu, K.}, \bibinfo{author}{Watanabe, S.}, \bibinfo{year}{2019}.
\newblock \bibinfo{title}{End-to-end neural speaker diarization with self-attention}, in: \bibinfo{booktitle}{IEEE Automatic Speech Recognition and Understanding Workshop}.
%Type = Inproceedings
\bibitem[{Gu and Dao(2024)}]{gu2024mamba}
\bibinfo{author}{Gu, A.}, \bibinfo{author}{Dao, T.}, \bibinfo{year}{2024}.
\newblock \bibinfo{title}{Mamba: Linear-time sequence modeling with selective state spaces}, in: \bibinfo{booktitle}{Conference on Language Modeling (COLM)}.
%Type = Inproceedings
\bibitem[{Gu et~al.(2022)Gu, Goel and R{\'e}}]{gu2022s4}
\bibinfo{author}{Gu, A.}, \bibinfo{author}{Goel, K.}, \bibinfo{author}{R{\'e}, C.}, \bibinfo{year}{2022}.
\newblock \bibinfo{title}{Efficiently modeling long sequences with structured state spaces}, in: \bibinfo{booktitle}{International Conference on Learning Representations}.
%Type = Inproceedings
\bibitem[{Hooker et~al.(2019)Hooker, Erhan, Kindermans and Kim}]{hooker2019benchmark}
\bibinfo{author}{Hooker, S.}, \bibinfo{author}{Erhan, D.}, \bibinfo{author}{Kindermans, P.J.}, \bibinfo{author}{Kim, B.}, \bibinfo{year}{2019}.
\newblock \bibinfo{title}{A benchmark for interpretability methods in deep neural networks}, in: \bibinfo{booktitle}{Advances in Neural Information Processing Systems}.
%Type = Inproceedings
\bibitem[{Horiguchi et~al.(2020)Horiguchi, Fujita, Watanabe, Xue and Nagamatsu}]{horiguchi2020eendeda}
\bibinfo{author}{Horiguchi, S.}, \bibinfo{author}{Fujita, Y.}, \bibinfo{author}{Watanabe, S.}, \bibinfo{author}{Xue, Y.}, \bibinfo{author}{Nagamatsu, K.}, \bibinfo{year}{2020}.
\newblock \bibinfo{title}{End-to-end speaker diarization for an unknown number of speakers with encoder-decoder based attractors}, in: \bibinfo{booktitle}{Interspeech}, pp. \bibinfo{pages}{269--273}.
%Type = Inproceedings
\bibitem[{Huang and Belongie(2017)}]{huang2017adain}
\bibinfo{author}{Huang, X.}, \bibinfo{author}{Belongie, S.J.}, \bibinfo{year}{2017}.
\newblock \bibinfo{title}{Arbitrary style transfer in real-time with adaptive instance normalization}, in: \bibinfo{booktitle}{IEEE International Conference on Computer Vision}, pp. \bibinfo{pages}{1510--1519}.
%Type = Article
\bibitem[{Hudson et~al.(1992)Hudson, Slatkin and Maddison}]{hudson1992estimation}
\bibinfo{author}{Hudson, R.R.}, \bibinfo{author}{Slatkin, M.}, \bibinfo{author}{Maddison, W.P.}, \bibinfo{year}{1992}.
\newblock \bibinfo{title}{Estimation of levels of gene flow from {DNA} sequence data}.
\newblock \bibinfo{journal}{Genetics} \bibinfo{volume}{132}, \bibinfo{pages}{583--589}.
%Type = Inproceedings
\bibitem[{Ioffe and Szegedy(2015)}]{ioffe2015batch}
\bibinfo{author}{Ioffe, S.}, \bibinfo{author}{Szegedy, C.}, \bibinfo{year}{2015}.
\newblock \bibinfo{title}{Batch normalization: Accelerating deep network training by reducing internal covariate shift}, in: \bibinfo{booktitle}{International Conference on Machine Learning}, pp. \bibinfo{pages}{448--456}.
%Type = Inproceedings
\bibitem[{Islam et~al.(2020)Islam, Jia and Bruce}]{islam2020position}
\bibinfo{author}{Islam, M.A.}, \bibinfo{author}{Jia, S.}, \bibinfo{author}{Bruce, N.D.B.}, \bibinfo{year}{2020}.
\newblock \bibinfo{title}{How much position information do convolutional neural networks encode?}, in: \bibinfo{booktitle}{International Conference on Learning Representations}.
%Type = Inproceedings
\bibitem[{Janner et~al.(2022)Janner, Du, Tenenbaum and Levine}]{janner2022planning}
\bibinfo{author}{Janner, M.}, \bibinfo{author}{Du, Y.}, \bibinfo{author}{Tenenbaum, J.B.}, \bibinfo{author}{Levine, S.}, \bibinfo{year}{2022}.
\newblock \bibinfo{title}{Planning with diffusion for flexible behavior synthesis}, in: \bibinfo{booktitle}{International Conference on Machine Learning}, pp. \bibinfo{pages}{9902--9915}.
%Type = Inproceedings
\bibitem[{Kayhan and van Gemert(2020)}]{kayhan2020translation}
\bibinfo{author}{Kayhan, O.S.}, \bibinfo{author}{van Gemert, J.C.}, \bibinfo{year}{2020}.
\newblock \bibinfo{title}{On translation invariance in cnns: Convolutional layers can exploit absolute spatial location}, in: \bibinfo{booktitle}{IEEE Conference on Computer Vision and Pattern Recognition}, pp. \bibinfo{pages}{14262--14273}.
%Type = Article
\bibitem[{Kelley et~al.(2018)Kelley, Reshef, Bileschi, Belanger, McLean and Snoek}]{kelley2018sequential}
\bibinfo{author}{Kelley, D.R.}, \bibinfo{author}{Reshef, Y.A.}, \bibinfo{author}{Bileschi, M.}, \bibinfo{author}{Belanger, D.}, \bibinfo{author}{McLean, C.Y.}, \bibinfo{author}{Snoek, J.}, \bibinfo{year}{2018}.
\newblock \bibinfo{title}{Sequential regulatory activity prediction across chromosomes with convolutional neural networks}.
\newblock \bibinfo{journal}{Genome Research} \bibinfo{volume}{28}, \bibinfo{pages}{739--750}.
%Type = Inproceedings
\bibitem[{Li and Janson(2024)}]{li2024optimal}
\bibinfo{author}{Li, M.}, \bibinfo{author}{Janson, L.}, \bibinfo{year}{2024}.
\newblock \bibinfo{title}{Optimal ablation for interpretability}, in: \bibinfo{booktitle}{Advances in Neural Information Processing Systems}.
%Type = Inproceedings
\bibitem[{Liu et~al.(2019)Liu, Sun, Zhou, Huang and Darrell}]{liu2019rethinking}
\bibinfo{author}{Liu, Z.}, \bibinfo{author}{Sun, M.}, \bibinfo{author}{Zhou, T.}, \bibinfo{author}{Huang, G.}, \bibinfo{author}{Darrell, T.}, \bibinfo{year}{2019}.
\newblock \bibinfo{title}{Rethinking the value of network pruning}, in: \bibinfo{booktitle}{International Conference on Learning Representations}.
%Type = Inproceedings
\bibitem[{Luo et~al.(2016)Luo, Li, Urtasun and Zemel}]{luo2016effective}
\bibinfo{author}{Luo, W.}, \bibinfo{author}{Li, Y.}, \bibinfo{author}{Urtasun, R.}, \bibinfo{author}{Zemel, R.}, \bibinfo{year}{2016}.
\newblock \bibinfo{title}{Understanding the effective receptive field in deep convolutional neural networks}, in: \bibinfo{booktitle}{Advances in Neural Information Processing Systems}.
%Type = Article
\bibitem[{Luo and Mesgarani(2019)}]{luo2019convtasnet}
\bibinfo{author}{Luo, Y.}, \bibinfo{author}{Mesgarani, N.}, \bibinfo{year}{2019}.
\newblock \bibinfo{title}{Conv-tasnet: Surpassing ideal time--frequency magnitude masking for speech separation}.
\newblock \bibinfo{journal}{IEEE/ACM Transactions on Audio, Speech, and Language Processing} \bibinfo{volume}{27}, \bibinfo{pages}{1256--1266}.
%Type = Article
\bibitem[{McGrath et~al.(2023)McGrath, Rahtz, Kramar, Mikulik and Legg}]{mcgrath2023hydra}
\bibinfo{author}{McGrath, T.}, \bibinfo{author}{Rahtz, M.}, \bibinfo{author}{Kramar, J.}, \bibinfo{author}{Mikulik, V.}, \bibinfo{author}{Legg, S.}, \bibinfo{year}{2023}.
\newblock \bibinfo{title}{The hydra effect: Emergent self-repair in language model computations}.
\newblock \bibinfo{journal}{arXiv} \DOIprefix\doi{10.48550/arXiv.2307.15771}. \bibinfo{note}{preprint}.
%Type = Article
\bibitem[{Meyes et~al.(2019)Meyes, Lu, de~Puiseau and Meisen}]{meyes2019ablation}
\bibinfo{author}{Meyes, R.}, \bibinfo{author}{Lu, M.}, \bibinfo{author}{de~Puiseau, C.W.}, \bibinfo{author}{Meisen, T.}, \bibinfo{year}{2019}.
\newblock \bibinfo{title}{Ablation studies in artificial neural networks}.
\newblock \bibinfo{journal}{arXiv} \DOIprefix\doi{10.48550/arXiv.1901.08644}. \bibinfo{note}{preprint}.
%Type = Inproceedings
\bibitem[{Morcos et~al.(2018)Morcos, Barrett, Rabinowitz and Botvinick}]{morcos2018single}
\bibinfo{author}{Morcos, A.S.}, \bibinfo{author}{Barrett, D.G.T.}, \bibinfo{author}{Rabinowitz, N.C.}, \bibinfo{author}{Botvinick, M.}, \bibinfo{year}{2018}.
\newblock \bibinfo{title}{On the importance of single directions for generalization}, in: \bibinfo{booktitle}{International Conference on Learning Representations}.
%Type = Inproceedings
\bibitem[{van~den Oord et~al.(2016)van~den Oord, Dieleman, Zen, Simonyan, Vinyals, Graves, Kalchbrenner, Senior and Kavukcuoglu}]{oord2016wavenet}
\bibinfo{author}{van~den Oord, A.}, \bibinfo{author}{Dieleman, S.}, \bibinfo{author}{Zen, H.}, \bibinfo{author}{Simonyan, K.}, \bibinfo{author}{Vinyals, O.}, \bibinfo{author}{Graves, A.}, \bibinfo{author}{Kalchbrenner, N.}, \bibinfo{author}{Senior, A.}, \bibinfo{author}{Kavukcuoglu, K.}, \bibinfo{year}{2016}.
\newblock \bibinfo{title}{{WaveNet}: A generative model for raw audio}, in: \bibinfo{booktitle}{9th ISCA Speech Synthesis Workshop (SSW 9)}, p. \bibinfo{pages}{125}.
%Type = Article
\bibitem[{Perslev et~al.(2021)Perslev, Darkner, Kempfner, Nikolic, Jennum and Igel}]{perslev2021usleep}
\bibinfo{author}{Perslev, M.}, \bibinfo{author}{Darkner, S.}, \bibinfo{author}{Kempfner, L.}, \bibinfo{author}{Nikolic, M.}, \bibinfo{author}{Jennum, P.J.}, \bibinfo{author}{Igel, C.}, \bibinfo{year}{2021}.
\newblock \bibinfo{title}{{U-S}leep: Resilient high-frequency sleep staging}.
\newblock \bibinfo{journal}{npj Digital Medicine} \bibinfo{volume}{4}, \bibinfo{pages}{72}.
%Type = Inproceedings
\bibitem[{Perslev et~al.(2019)Perslev, Jensen, Darkner, Jennum and Igel}]{perslev2019utime}
\bibinfo{author}{Perslev, M.}, \bibinfo{author}{Jensen, M.H.}, \bibinfo{author}{Darkner, S.}, \bibinfo{author}{Jennum, P.J.}, \bibinfo{author}{Igel, C.}, \bibinfo{year}{2019}.
\newblock \bibinfo{title}{{U-Time}: A fully convolutional network for time series segmentation applied to sleep staging}, in: \bibinfo{booktitle}{Advances in Neural Information Processing Systems}, pp. \bibinfo{pages}{4417--4428}.
%Type = Article
\bibitem[{Pfrommer et~al.(2025)Pfrommer, Ma, Huang and Sojoudi}]{pfrommer2025spooky}
\bibinfo{author}{Pfrommer, S.}, \bibinfo{author}{Ma, G.}, \bibinfo{author}{Huang, Y.}, \bibinfo{author}{Sojoudi, S.}, \bibinfo{year}{2025}.
\newblock \bibinfo{title}{Spooky action at a distance: Normalization layers enable side-channel spatial communication}.
\newblock \bibinfo{journal}{arXiv} \DOIprefix\doi{10.48550/arXiv.2507.04709}. \bibinfo{note}{preprint}.
%Type = Article
\bibitem[{Phan et~al.(2022)Phan, Mikkelsen, Ch{\'e}n, Koch, Mertins and De~Vos}]{phan2022sleeptransformer}
\bibinfo{author}{Phan, H.}, \bibinfo{author}{Mikkelsen, K.}, \bibinfo{author}{Ch{\'e}n, O.Y.}, \bibinfo{author}{Koch, P.}, \bibinfo{author}{Mertins, A.}, \bibinfo{author}{De~Vos, M.}, \bibinfo{year}{2022}.
\newblock \bibinfo{title}{{SleepTransformer}: Automatic sleep staging with interpretability and uncertainty quantification}.
\newblock \bibinfo{journal}{IEEE Transactions on Biomedical Engineering} \bibinfo{volume}{69}, \bibinfo{pages}{2456--2467}.
%Type = Inproceedings
\bibitem[{Ronneberger et~al.(2015)Ronneberger, Fischer and Brox}]{ronneberger2015unet}
\bibinfo{author}{Ronneberger, O.}, \bibinfo{author}{Fischer, P.}, \bibinfo{author}{Brox, T.}, \bibinfo{year}{2015}.
\newblock \bibinfo{title}{U-net: Convolutional networks for biomedical image segmentation}, in: \bibinfo{booktitle}{Medical Image Computing and Computer-Assisted Intervention}.
%Type = Inproceedings
\bibitem[{Schneider et~al.(2020)Schneider, Rusak, Eck, Bringmann, Brendel and Bethge}]{schneider2020improving}
\bibinfo{author}{Schneider, S.}, \bibinfo{author}{Rusak, E.}, \bibinfo{author}{Eck, L.}, \bibinfo{author}{Bringmann, O.}, \bibinfo{author}{Brendel, W.}, \bibinfo{author}{Bethge, M.}, \bibinfo{year}{2020}.
\newblock \bibinfo{title}{Improving robustness against common corruptions by covariate shift adaptation}, in: \bibinfo{booktitle}{Advances in Neural Information Processing Systems}.
%Type = Article
\bibitem[{Supratak et~al.(2017)Supratak, Dong, Wu and Guo}]{supratak2017deepsleepnet}
\bibinfo{author}{Supratak, A.}, \bibinfo{author}{Dong, H.}, \bibinfo{author}{Wu, C.}, \bibinfo{author}{Guo, Y.}, \bibinfo{year}{2017}.
\newblock \bibinfo{title}{{DeepSleepNet}: A model for automatic sleep stage scoring based on raw single-channel {EEG}}.
\newblock \bibinfo{journal}{IEEE Transactions on Neural Systems and Rehabilitation Engineering} \bibinfo{volume}{25}, \bibinfo{pages}{1998--2008}.
%Type = Inproceedings
\bibitem[{Supratak and Guo(2020)}]{supratak2020tinysleepnet}
\bibinfo{author}{Supratak, A.}, \bibinfo{author}{Guo, Y.}, \bibinfo{year}{2020}.
\newblock \bibinfo{title}{{TinySleepNet}: An efficient deep learning model for sleep stage scoring based on raw single-channel {EEG}}, in: \bibinfo{booktitle}{International Conference of the IEEE Engineering in Medicine and Biology Society (EMBC)}, pp. \bibinfo{pages}{641--644}.
%Type = Article
\bibitem[{{The 1000 Genomes Project Consortium}(2015)}]{tgp2015global}
\bibinfo{author}{{The 1000 Genomes Project Consortium}}, \bibinfo{year}{2015}.
\newblock \bibinfo{title}{A global reference for human genetic variation}.
\newblock \bibinfo{journal}{Nature} \bibinfo{volume}{526}, \bibinfo{pages}{68--74}.
%Type = Article
\bibitem[{Tian(2026)}]{tian2026lai}
\bibinfo{author}{Tian, Q.}, \bibinfo{year}{2026}.
\newblock \bibinfo{title}{What limits local ancestry inference at low divergence: a feasibility threshold, a metric that conceals failure, and a deficit of input more than architecture}.
\newblock \bibinfo{journal}{bioRxiv} \URLprefix \url{https://doi.org/10.64898/2026.07.30.741148}, \DOIprefix\doi{10.64898/2026.07.30.741148}. \bibinfo{note}{preprint}.
%Type = Article
\bibitem[{Tian et~al.(2021)Tian, Arbel and Clark}]{tian2021lda}
\bibinfo{author}{Tian, Q.}, \bibinfo{author}{Arbel, T.}, \bibinfo{author}{Clark, J.J.}, \bibinfo{year}{2021}.
\newblock \bibinfo{title}{Task dependent deep {LDA} pruning of neural networks}.
\newblock \bibinfo{journal}{Computer Vision and Image Understanding} \bibinfo{volume}{203}, \bibinfo{pages}{103154}.
%Type = Article
\bibitem[{Ulyanov et~al.(2016)Ulyanov, Vedaldi and Lempitsky}]{ulyanov2016instance}
\bibinfo{author}{Ulyanov, D.}, \bibinfo{author}{Vedaldi, A.}, \bibinfo{author}{Lempitsky, V.}, \bibinfo{year}{2016}.
\newblock \bibinfo{title}{Instance normalization: The missing ingredient for fast stylization}.
\newblock \bibinfo{journal}{arXiv} \DOIprefix\doi{10.48550/arXiv.1607.08022}. \bibinfo{note}{preprint}.
%Type = Inproceedings
\bibitem[{Vaswani et~al.(2017)Vaswani, Shazeer, Parmar, Uszkoreit, Jones, Gomez, Kaiser and Polosukhin}]{vaswani2017attention}
\bibinfo{author}{Vaswani, A.}, \bibinfo{author}{Shazeer, N.}, \bibinfo{author}{Parmar, N.}, \bibinfo{author}{Uszkoreit, J.}, \bibinfo{author}{Jones, L.}, \bibinfo{author}{Gomez, A.N.}, \bibinfo{author}{Kaiser, L.}, \bibinfo{author}{Polosukhin, I.}, \bibinfo{year}{2017}.
\newblock \bibinfo{title}{Attention is all you need}, in: \bibinfo{booktitle}{Advances in Neural Information Processing Systems}.
%Type = Inproceedings
\bibitem[{Wang et~al.(2021)Wang, Shelhamer, Liu, Olshausen and Darrell}]{wang2021tent}
\bibinfo{author}{Wang, D.}, \bibinfo{author}{Shelhamer, E.}, \bibinfo{author}{Liu, S.}, \bibinfo{author}{Olshausen, B.}, \bibinfo{author}{Darrell, T.}, \bibinfo{year}{2021}.
\newblock \bibinfo{title}{Tent: Fully test-time adaptation by entropy minimization}, in: \bibinfo{booktitle}{International Conference on Learning Representations}.
%Type = Inproceedings
\bibitem[{Wu and He(2018)}]{wu2018group}
\bibinfo{author}{Wu, Y.}, \bibinfo{author}{He, K.}, \bibinfo{year}{2018}.
\newblock \bibinfo{title}{Group normalization}, in: \bibinfo{booktitle}{European Conference on Computer Vision}, pp. \bibinfo{pages}{3--19}.
%Type = Article
\bibitem[{Wu and Johnson(2021)}]{wu2021rethinking}
\bibinfo{author}{Wu, Y.}, \bibinfo{author}{Johnson, J.}, \bibinfo{year}{2021}.
\newblock \bibinfo{title}{Rethinking ``batch'' in batchnorm}.
\newblock \bibinfo{journal}{arXiv} \DOIprefix\doi{10.48550/arXiv.2105.07576}. \bibinfo{note}{preprint}.
%Type = Inproceedings
\bibitem[{Yi et~al.(2021)Yi, Wen and Jiang}]{yi2021asformer}
\bibinfo{author}{Yi, F.}, \bibinfo{author}{Wen, H.}, \bibinfo{author}{Jiang, T.}, \bibinfo{year}{2021}.
\newblock \bibinfo{title}{{ASF}ormer: Transformer for action segmentation}, in: \bibinfo{booktitle}{British Machine Vision Conference}.
%Type = Inproceedings
\bibitem[{Yu and Koltun(2016)}]{yu2016dilated}
\bibinfo{author}{Yu, F.}, \bibinfo{author}{Koltun, V.}, \bibinfo{year}{2016}.
\newblock \bibinfo{title}{Multi-scale context aggregation by dilated convolutions}, in: \bibinfo{booktitle}{International Conference on Learning Representations}.

\end{thebibliography}

\appendix

% Elsevier numbers appendix material by the letter of the appendix it sits in:
% Table A.1 in Appendix A, Table B.1 in Appendix B, equations likewise.
% \numberwithin ties each counter to the section and resets it at every
% appendix, which produces that scheme; it also keeps the letter at every
% cross-reference, which is what the old manual "A" prefix was for.
\numberwithin{table}{section}
\numberwithin{figure}{section}
\numberwithin{equation}{section}
% \thesection expands to "Appendix~A" in this class, which would render as
% "Table Appendix~A.1"; the counter letter alone is what Elsevier prints.
\renewcommand{\thetable}{\Alph{section}.\arabic{table}}
\renewcommand{\thefigure}{\Alph{section}.\arabic{figure}}
\renewcommand{\theequation}{\Alph{section}.\arabic{equation}}

\section{Architectures and training}
\label{app:arch}

Enough to reimplement, and in particular enough to check the two properties the comparisons rest on: that the per-position control differs from GroupNorm in the pooling axes and in nothing else, and that a $k$-block prefix is exactly the full stack with the later blocks absent rather than a separately designed smaller model.

\paragraph{The dilated stack} Stem: $\mathrm{Conv1d}(C \to 64,\ k{=}5)$. Each of nine blocks is $\mathrm{norm} \to \mathrm{GELU} \to \mathrm{Conv1d}(64 \to 64,\ k{=}5,\ \mathrm{dilation}{=}d) \to \mathrm{norm} \to \mathrm{GELU} \to \mathrm{Conv1d}(64 \to 64,\ k{=}1)$, added residually, with $d = 1, 2, \dots, 256$. Head: $\mathrm{norm} \to \mathrm{GELU} \to \mathrm{Conv1d}(64 \to 1,\ k{=}1)$. A kernel-5 convolution at dilation $d$ adds $4d$ positions, so the prefix of the first $k$ blocks has receptive field $1 + 4\sum_{i<k} 2^{i} + 4$: \reachMin, 33, 129, 513 and \reachMax\ at $k = 1, 3, 5, 7, 9$. All nine blocks give \nParams\ parameters, one block \nParamsLo.

\paragraph{Normalization variants} GroupNorm uses 8 groups. The per-position control normalizes over exactly the same 8 channel groups with the statistics taken at each position independently, and carries the same affine parameters, so the pooling axes are the only difference between them. InstanceNorm pools over length within a channel; BatchNorm over batch and length while training and over running statistics at evaluation. Conv-TasNet's gLN is a single group over channels and time, and its cLN the causal counterpart, statistics over channels and time up to $t$ by a cumulative sum.

\paragraph{The Conv-TasNet separator} As published: a $1\times1$ bottleneck of 64 channels expanding to 184, a depthwise convolution of kernel 3 at dilation $d$, and a $1\times1$ projection back, with normalization after each PReLU and the block added residually. \tasParams\ parameters. Its depthwise kernel is 3, so a block adds $2d$ rather than $4d$, and its receptive fields are \tasReachLo, 19, 67, 259 and \tasReachHi.

\paragraph{The transformer} Nine pre-norm blocks, width 64, four heads, a pointwise feed-forward of the same width, and a fixed sinusoidal positional encoding on the stem, without which attention is permutation-equivariant and the model could not tell a switch at position 10 from one at position 3000. Attention runs on tokens mean-pooled 16-fold and is upsampled back: full \window-position attention is 16M entries per head per sequence and exceeds what the accelerator will allocate. Label runs span hundreds of positions, so the pooling discards little, but it does cap the resolution at which attention can localize a switch and is reported for that reason.

\paragraph{The U-Net} Stem: $\mathrm{Conv1d}(C \to 64,\ k{=}5)$. A block is $\mathrm{norm} \to \mathrm{GELU} \to \mathrm{Conv1d}(64 \to 64,\ k{=}5)$ throughout. The encoder applies one block per level and halves the sequence with $\mathrm{AvgPool1d}(2)$; one block runs at the bottleneck; the decoder upsamples by nearest-neighbor interpolation to the matching encoder length, concatenates that level's skip, projects the doubled width back with $\mathrm{Conv1d}(128 \to 64,\ k{=}1)$, and applies one block. Head: $\mathrm{norm} \to \mathrm{GELU} \to \mathrm{Conv1d}(64 \to 1,\ k{=}1)$. Depth runs from 1 to \unetDepths, which is what sets reach, and takes the parameter count from \unetParamsLo\ to \unetParamsHi; depth \unetMatchDepth\ at \unetMatchParams\ is the closest to the dilated stack's \nParams, within \unetMatchPct\%. Width is held at 64 as everywhere else, so depth is the only free variable.

Two things differ from the other architectures and both are deliberate. The receptive field is measured rather than derived: a recursion through down- and up-sampling is easy to get wrong by one, and the whole experiment hinges on that number (Section~\ref{sec:bounds}). And the blocks are not residual, so removing one is not the identity map and this architecture has no block-ablation story: the model raises rather than silently accepting a \texttt{skip} argument, since a U-Net ablated the way Section~\ref{sec:ablation} ablates the dilated stack would not be measuring the same thing.

\paragraph{Training} AdamW, learning rate $10^{-3}$, weight decay $10^{-4}$, cosine schedule, batch 32, 15 epochs, binary cross-entropy on logits. Held-out accuracy plateaus by epoch three at these settings; 15 is kept so that no comparison between normalizations is a comparison of who converges fastest. The seed is set before construction as well as before training, so initialization and batch order are both matched across the variants being compared. Widths are held at 64 everywhere, and every convolution is zero-padded identically in every variant, so the positional information padding injects \citep{islam2020position} is the same in every row of every comparison and cancels from the differences we report.

\section{The three processes}
\label{app:data}

All three must supply exact per-position labels and a difficulty that is set rather than estimated, since without the first there is nothing to measure against and without the second there is no axis to sweep. What they do not share is how they get there (a coalescent, a Gaussian emission model, and a public genome), which is what makes their agreement evidence rather than repetition.

\paragraph{Simulated ancestry} Two populations split $T$ generations ago from an ancestor of effective size $10^4$ and are simulated over $10^7$ bases at recombination rate $10^{-8}$ and mutation rate $1.25 \times 10^{-8}$ per base per generation. Each population contributes 100 reference and 100 donor haplotypes. Donors are spliced into 64 mosaic sequences per replicate, with breakpoints from a Poisson process at rate $g \rho L$ for $g$ generations since admixture: 30 unless the switch rate is being varied, which is the only thing $g$ changes. Four channels per position carry the observed allele, the two reference allele frequencies, and their log likelihood ratio; frequencies are estimated from the cropped sequence rather than the whole simulation, so they describe exactly the sites the model sees. Sequences of \window\ positions are cut at random offsets, since segments are long relative to the sequence and a fixed offset would give the class balance a positional bias. The per-position evidence weakens as the populations' allele frequencies converge, which is what the divergence sweep of Section~\ref{sec:tests} varies, at \factLevels\ levels; measured $F_{ST}$ matches the benchmark the demography was ported from to within \portMaxDiff\ absolute at all \portLevels\ split times of the port check (at most \portMaxRel\% relative, at the hardest level, where $F_{ST}$ is \portFstLo; Table~\ref{tab:port}).

Training and evaluation sets come from different coalescent simulations. Sequences within one replicate share reference panels, so a split inside a single draw scores memorization of those panels rather than generalization; an earlier version of the port did exactly that and read below chance at levels the benchmark solves at 0.89.

\paragraph{Gaussian--Markov} Labels from a symmetric two-state Markov chain with the switch probability set directly; emissions $x_t \sim \mathcal{N}(\pm m/2,\ I)$ in four channels with $\|m\| = \delta$, the signal spread evenly across channels rather than concentrated in one, since a single informative channel would let a network solve the task without mixing channels at all.

Spread this way, the statistic a single-group normalization layer pools is not merely correlated with $\bar\pi$ but is, up to noise, a known affine function of it. Summing the emission model over the $C$ channels and the $L$-position sequence,
\begin{equation}
\mu_S \;=\; \frac{\delta}{2\sqrt{C}}\,(2\bar\pi - 1) \;+\; \eta,
\qquad \eta \sim \mathcal{N}\!\big(0,\ 1/(CL)\big),
\label{eq:signal}
\end{equation}
since the noise terms are independent standard normals averaged over $CL$ values and the signal term is uniform across channels by construction; nothing else contributes. We confirm this against the raw draws at every switch density used in the dense sweep of Section~\ref{sec:tests}: over \sigCells\ cells the ratio of the measured residual's standard deviation to $1/\sqrt{CL}$ is \sigRatioLo--\sigRatioHi, and the correlation between $\mu_S$ and \Eqref{eq:signal}'s prediction falls from \sigCorrHi\ where labels rarely switch to \sigCorrLo\ at the densest setting, which is the correct direction, since the signal term's own variance shrinks as $\bar\pi$ concentrates near one half while the noise floor does not. This is the sense in which \emph{the summary alone} is not an analogy for what the path supplies on this process, but the quantity itself, plus a noise floor that shrinks with sequence length and is otherwise unrelated to $\bar\pi$.

We fix $\delta = \synDelta$. That value is chosen so that the accuracy achievable at each reach matches the ancestry process, not so that the per-position error is a round number: at $\delta = \synDelta$ one position gives \perPosBayes, nine give \closedNine\ and 2049 give \closedFar, against the ancestry process's \genLoAcc\ and \genHiAcc\ at the two ends. Choosing $\delta$ by per-position error instead put nine positions at 0.73 and left nothing for reach to buy.

\paragraph{Real haplotypes} 1000 Genomes chromosome 22, 16--51\,Mb, CEU (Utah residents of northern and western European ancestry) and GIH (Gujarati Indians in Houston), 129{,}828 biallelic SNVs. Reference and donor haplotypes are disjoint partitions of each population's panel, 80 each, so a donor is never in the panel used to classify it. Mosaics and features are built by the same code as the simulated process. Training sequences come from the first 60\% of sites and evaluation sequences from the last 35\%, with a 5\% buffer between them, because linkage disequilibrium correlates nearby sequences and adjacent segments would leak. The three replicates are independent reference/donor partitions and independent mosaics rather than independent draws from a generating process, which is weaker, and is the reason the real-data arm is reported with that caveat in Section~\ref{sec:discussion}.

\section{Numerical check of \Eqref{eq:jacobian}}
\label{app:bounds}

The identity is what makes the exposure criterion checkable against a layer's definition instead of merely assertible, so it is verified numerically rather than trusted. The Jacobian is $|S| \times |S|$, so it can be built in full only on a small input; the identity does not depend on the size. The layer compared against the closed form is instantiated with $\epsilon = 0$, since the identity is stated for the unstabilized layer (Section~\ref{sec:theory}); cLN alone carries $\epsilon = 10^{-12}$, because its cumulative variance is exactly zero at the first position. On $\jacC$ channels and $\jacL$ positions, $|S| = \jacS$:

\begin{center}
\small
\begin{tabular}{lr}
\toprule
largest disagreement with \Eqref{eq:jacobian}, gLN & $\jacErr$ \\
mean $|\partial z_t / \partial x_s|$, $s \neq t$ & $\jacOffdiag$ \\
$1/|S|$ & $\jacInvS$ \\
$\partial z_t / \partial x_t$ & $\jacDiag$ \\
$\sum_{s \neq t} \partial z_t / \partial x_s$ & $\jacRowsum$ \\
row total & $\jacRowTotal$ \\
\midrule
BatchNorm at evaluation, largest $|\partial z_t/\partial x_s|$, $s \neq t$ & $0$ \\
cLN, largest $|\partial z_t/\partial x_s|$ for $s > t$ & $0$ \\
cLN, largest $|\partial z_t/\partial x_s|$ for $s < t$ & $\jacClnPast$ \\
\bottomrule
\end{tabular}
\end{center}

The off-diagonal terms are of order $1/|S|$ and sum to minus the diagonal, so the row total is zero to machine precision. BatchNorm's off-diagonal terms at evaluation and cLN's future terms are not small but exactly zero, which is what makes the two predictions of absence in (P4) predictions rather than expectations.

\section{A premise that did not survive its own control}
\label{app:deff}

Everything in this paper rests on the per-position variant genuinely closing the path rather than appearing to, which is why \ref{app:bounds} checks that it does. That discipline came from a failure, which we record because the way it failed is a general hazard.

The project began from a different observation on the same benchmark. The effective discriminant dimensionality of the final representation (the participation ratio of the per-channel Fisher ratios \citep{tian2021lda}, a scale-free count of how widely class information is spread across channels) appeared non-monotonic in task difficulty: \deffObsEasy\ at the easy end, rising to \deffObsPeak\ at intermediate divergence, collapsing to \deffObsHard\ at the floor. A representation that spreads class information more widely as the task gets harder and then loses it is an appealing story.

It did not survive a label null. Recomputing the same statistic on the same activations, with labels drawn from an independent replicate at the same difficulty (a second draw from the same process, statistically identical to the real labels and causally unrelated to those activations) gives \deffNullEasy\ at the easy end falling to \deffNullHard\ at the floor. The null moves along the axis in the same direction as the observation, and most of the apparent collapse is the null moving. Across all \deffLevels\ levels, the observed value exceeds its null by at most \deffMarginMax\ of the null's own spread, so there is nothing there.

The first null we tried was worse than uninformative. It rolled each sequence's label vector circularly, which preserves the run-length structure while destroying the alignment between labels and activations, except that a circular shift is exactly the identity on a sequence whose label never changes, and \shiftSingle\% of these sequences are like that. Measured on the same label vectors, a shift leaves precisely the single-label sequences bit-identical and no others, and alters \shiftChanged\% of positions in total. It agreed closely with the observed value, which looked alarming and meant nothing.

Two things carried into the present work. A scale-free statistic needs a null computed on the same activations, and that null cannot be assumed constant across conditions: the difficulty axis has a known floor precisely so that a representation which has run out of information can be told from one that is merely struggling. And a control has to be verified to remove what it claims to remove, which is why \ref{app:bounds} establishes that BatchNorm's off-diagonal terms at evaluation and cLN's future terms are not merely small but exactly zero, and why the per-position control is matched to GroupNorm group for group so that the pooling axes are the only thing that differs.

\section{Cost}
\label{app:cost}

The experiments are \numExperiments\ trained-and-evaluated configurations in total, each a network trained from scratch for 15 epochs on \trainSeq\ sequences of \window\ positions and evaluated on \evalSeq. They ran on a single consumer GPU. We do not report wall-clock: the machine slept between cells for part of the sweep, so the recorded durations include time not computing and would misstate the cost in either direction depending on the cell. The configuration count is exact and is the honest unit here.

\section{Full results}
\label{app:results}

In order: Table~\ref{tab:grid} is the accuracy grid behind (P1) and Table~\ref{tab:difficulty} the reach worth it implies at each divergence level; Table~\ref{tab:dose} is the dose--response behind (P3), on both simulated processes and on real haplotypes; Table~\ref{tab:tasnet} gives the Conv-TasNet curves and Table~\ref{tab:transformer} the transformer, the two architecture arms; Table~\ref{tab:resid} places the per-position networks against the local bound for their own receptive field, cell by cell; Table~\ref{tab:optima} lists the local bounds and exactly computed optima and Table~\ref{tab:port} the simulated process against the benchmark it was ported from; Table~\ref{tab:unet} is the U-Net depth sweep; Table~\ref{tab:survey} is the normalization survey of published models discussed in Section~\ref{sec:discussion}; Table~\ref{tab:ablcost} gives the per-level costs behind the pooled ablation numbers of Section~\ref{sec:ablation}; and Table~\ref{tab:deriv} sets \Eqref{eq:advantage} against the optima it predicts.

% Generated by paper/make_numbers.py from results/*.json.
% Do not edit by hand: the next run overwrites this file.
\begin{table}[t]
\centering
\caption{Held-out accuracy of networks retrained with the first $k$ blocks, by divergence level and normalization. The columns are receptive fields in positions. Means over \nSeeds\ seeds; this is the grid behind Figure~\ref{fig:reach}a and Table~\ref{tab:difficulty}.}
\label{tab:grid}
\resizebox{\ifdim\width>\linewidth\linewidth\else\width\fi}{!}{%
\begin{tabular}{llrrrrr}
\toprule
normalization & $F_{ST}$ & $R=9$ & $R=33$ & $R=129$ & $R=513$ & $R=2049$ \\
\midrule
pooled over length & 0.2405 & 0.9631 & 0.9817 & 0.9962 & 0.9982 & 0.9983 \\
 & 0.0719 & 0.9353 & 0.9478 & 0.9686 & 0.9836 & 0.9877 \\
 & 0.0380 & 0.9158 & 0.9250 & 0.9381 & 0.9532 & 0.9610 \\
 & 0.0199 & 0.8502 & 0.8649 & 0.8745 & 0.8907 & 0.8969 \\
 & 0.0096 & 0.7272 & 0.7339 & 0.7399 & 0.7507 & 0.7592 \\
per position & 0.2405 & 0.7228 & 0.8786 & 0.9849 & 0.9980 & 0.9983 \\
 & 0.0719 & 0.6204 & 0.7187 & 0.8618 & 0.9674 & 0.9862 \\
 & 0.0380 & 0.5767 & 0.6463 & 0.7644 & 0.8914 & 0.9508 \\
 & 0.0199 & 0.5521 & 0.5977 & 0.6765 & 0.7763 & 0.8720 \\
 & 0.0096 & 0.5190 & 0.5487 & 0.5945 & 0.6509 & 0.7200 \\
\bottomrule
\end{tabular}}
\end{table}

\begin{table}[t]
\centering
\caption{Reach worth at each divergence level of the simulated genomes, with statistics pooled along the sequence and per position. Their ratio, per position over pooled, does not vary systematically with difficulty.}
\label{tab:difficulty}
\resizebox{\ifdim\width>\linewidth\linewidth\else\width\fi}{!}{%
\begin{tabular}{lrrr}
\toprule
$F_{ST}$ & pooled over length & per position & ratio \\
\midrule
0.2405 & $+0.0351 \pm 0.0019$ & $+0.2754 \pm 0.0099$ & 7.8 \\
0.0719 & $+0.0524 \pm 0.0110$ & $+0.3658 \pm 0.0051$ & 7.0 \\
0.0380 & $+0.0452 \pm 0.0107$ & $+0.3741 \pm 0.0052$ & 8.3 \\
0.0199 & $+0.0467 \pm 0.0077$ & $+0.3199 \pm 0.0126$ & 6.8 \\
0.0096 & $+0.0320 \pm 0.0119$ & $+0.2011 \pm 0.0173$ & 6.3 \\
\bottomrule
\end{tabular}}
\end{table}

\begin{table}[t]
\centering
\caption{Switch-rate dose--response on both simulated processes and on real haplotypes. Both middle columns are the accuracy gained by the \reachFold-fold enlargement of the receptive field; the last column is the constant of the $c/n$ regularity at each density. It is flat within each process and differs between them.}
\label{tab:dose}
\resizebox{\ifdim\width>\linewidth\linewidth\else\width\fi}{!}{%
\begin{tabular}{llccrr}
\toprule
process & $n$ & pooled over length & per position & ratio & $(\text{ratio}-1)\,n$ \\
\midrule
genomic & 0.29 & $+0.0452 \pm 0.0107$ & $+0.3741 \pm 0.0052$ & 8.3 & 2.08 \\
 & 0.99 & $+0.1098 \pm 0.0070$ & $+0.3453 \pm 0.0074$ & 3.1 & 2.13 \\
 & 3.00 & $+0.1745 \pm 0.0026$ & $+0.2958 \pm 0.0061$ & 1.7 & 2.08 \\
 & 9.89 & $+0.1692 \pm 0.0089$ & $+0.2151 \pm 0.0062$ & 1.3 & 2.68 \\
synthetic & 0.09 & $+0.0062 \pm 0.0014$ & $+0.3892 \pm 0.0038$ & 62.8 & 5.80 \\
 & 0.29 & $+0.0246 \pm 0.0040$ & $+0.3740 \pm 0.0047$ & 15.2 & 4.05 \\
 & 0.99 & $+0.0785 \pm 0.0078$ & $+0.3349 \pm 0.0034$ & 4.3 & 3.25 \\
 & 2.98 & $+0.1245 \pm 0.0039$ & $+0.2630 \pm 0.0083$ & 2.1 & 3.32 \\
 & 5.99 & $+0.1269 \pm 0.0071$ & $+0.1990 \pm 0.0036$ & 1.6 & 3.40 \\
 & 9.89 & $+0.1094 \pm 0.0008$ & $+0.1543 \pm 0.0043$ & 1.4 & 4.05 \\
 & 19.87 & $+0.0860 \pm 0.0041$ & $+0.1022 \pm 0.0003$ & 1.2 & 3.74 \\
real 1000G & 0.13 & $+0.0181 \pm 0.0105$ & $+0.2190 \pm 0.0195$ & 12.1 & 1.40 \\
\bottomrule
\end{tabular}}
\end{table}

\begin{table}[t]
\centering
\caption{The Conv-TasNet separator on the synthetic process at 0.27 switches per sequence, against an exact whole-sequence optimum of 0.9522. Its depthwise kernel is 3, so a block adds $2d$ rather than $4d$ and the receptive fields differ from the other tables. Means over \nSeeds\ seeds.}
\label{tab:tasnet}
\resizebox{\ifdim\width>\linewidth\linewidth\else\width\fi}{!}{%
\begin{tabular}{llrrrrr}
\toprule
layer & statistics pooled over & $R=7$ & $R=19$ & $R=67$ & $R=259$ & $R=1027$ \\
\midrule
gLN & channels and time & 0.9246 & 0.9244 & 0.9245 & 0.9289 & 0.9407 \\
cLN & channels and time up to $t$ & 0.8676 & 0.8705 & 0.8740 & 0.8860 & 0.9186 \\
per position & channels & 0.5423 & 0.5681 & 0.6262 & 0.7314 & 0.8745 \\
\bottomrule
\end{tabular}}
\end{table}

\begin{table}[t]
\centering
\caption{(P4) A transformer, whose attention spans the input sequence at every depth, with sequence-pooled and with per-position normalization. The two are within a few thousandths of each other and of the exact optimum at every switch density: with the context already available there is nothing for the path to supply. The last column is the better of the two normalizations against the optimum, and the range the main text quotes is the widest of these. Means over \nSeeds\ seeds.}
\label{tab:transformer}
\resizebox{\ifdim\width>\linewidth\linewidth\else\width\fi}{!}{%
\begin{tabular}{rrrrrr}
\toprule
switches per sequence & pooled over length & per position & difference & exact optimum & shortfall \\
\midrule
0.27 & 0.9497 & 0.9472 & +0.0025 & 0.9522 & 0.0025 \\
0.98 & 0.8956 & 0.8932 & +0.0024 & 0.8993 & 0.0037 \\
2.98 & 0.8146 & 0.8125 & +0.0021 & 0.8194 & 0.0047 \\
9.79 & 0.7023 & 0.7000 & +0.0022 & 0.7058 & 0.0035 \\
\bottomrule
\end{tabular}}
\end{table}

\begin{table}[t]
\centering
\caption{Per-position networks against the local bound for their own receptive field, on the synthetic process: each entry is held-out accuracy minus that bound, so zero is the bound and positive is above it. This is the grid behind every number quoted in Section~\ref{sec:tests} for this comparison: \boundAllCells\ cells, mean absolute value \boundAllMean, \boundAllAbove\ of them positive, the largest \boundAllAboveMax. Exceedances are expected, since $\mathrm{local}(R)$ is a blocked bound and therefore conservative, and the entries are finite-sample means (Section~\ref{sec:setup}). Adding the corresponding entry of Table~\ref{tab:optima} recovers the accuracy itself. Means over \nSeeds\ seeds.}
\label{tab:resid}
\resizebox{\ifdim\width>\linewidth\linewidth\else\width\fi}{!}{%
\begin{tabular}{rrrrrr}
\toprule
switches/sequence & $R=9$ & $R=33$ & $R=129$ & $R=513$ & $R=2049$ \\
\midrule
0.09 & -0.0006 & -0.0017 & -0.0050 & -0.0124 & -0.0172 \\
0.29 & -0.0003 & -0.0016 & -0.0049 & -0.0115 & -0.0119 \\
0.99 & -0.0008 & -0.0016 & -0.0030 & -0.0049 & -0.0013 \\
2.98 & -0.0010 & -0.0007 & +0.0012 & +0.0019 & -0.0001 \\
5.99 & -0.0008 & +0.0005 & +0.0031 & +0.0084 & -0.0009 \\
9.89 & -0.0003 & +0.0010 & +0.0051 & +0.0091 & -0.0002 \\
19.87 & -0.0005 & +0.0011 & +0.0055 & +0.0045 & -0.0005 \\
\bottomrule
\end{tabular}}
\end{table}

\begin{table}[t]
\centering
\caption{The reference values on the synthetic process: the local bound at every receptive field it was computed for, and the exactly computed optima with full context and with an oracle given only the sequence's class proportion. The intermediate columns are here because the main text reads values off them: $R=33$ brackets the U-Net's receptive field (Section~\ref{sec:tests}) and $R=129$ is where the per-position control matches a sequence-pooled model at $R=9$ (Section~\ref{sec:tests}). At \optimaCrossN\ switches the proportion oracle exceeds full context by \optimaCross. The two references are not nested and neither bounds the other: the oracle is handed the sequence's class proportion, a function of the labels, while full context is the best any predictor of the observations can do. Where nearly every sequence carries one label, the oracle is right at every position of those sequences and a predictor reading evidence of strength $\delta$ is not, so the two cross.}
\label{tab:optima}
\resizebox{\ifdim\width>\linewidth\linewidth\else\width\fi}{!}{%
\begin{tabular}{rrrrrrrr}
\toprule
switches & $R=9$ & $R=33$ & $R=129$ & $R=513$ & $R=2049$ & full & proportion oracle \\
\midrule
0.09 & 0.5479 & 0.5908 & 0.6741 & 0.8165 & 0.9537 & 0.9759 & 0.9760 \\
0.29 & 0.5479 & 0.5905 & 0.6738 & 0.8110 & 0.9335 & 0.9504 & 0.9348 \\
0.99 & 0.5482 & 0.5903 & 0.6708 & 0.7951 & 0.8835 & 0.8986 & 0.8390 \\
2.98 & 0.5493 & 0.5913 & 0.6660 & 0.7666 & 0.8115 & 0.8194 & 0.7239 \\
5.99 & 0.5474 & 0.5875 & 0.6543 & 0.7211 & 0.7466 & 0.7508 & 0.6567 \\
9.89 & 0.5466 & 0.5846 & 0.6424 & 0.6869 & 0.7008 & 0.7031 & 0.6290 \\
19.87 & 0.5470 & 0.5821 & 0.6244 & 0.6439 & 0.6492 & 0.6508 & 0.5858 \\
\bottomrule
\end{tabular}}
\end{table}

\begin{table}[t]
\centering
\caption{The simulated ancestry process against the benchmark it was ported from \citep{tian2026lai}, at all eight of its split times. $F_{ST}$ is measured from the simulated panels rather than set, so agreement is a check on the demography, the coalescent draw and the estimator together. Accuracy is the reference architecture's, over \nSeeds\ seeds.}
\label{tab:port}
\resizebox{\ifdim\width>\linewidth\linewidth\else\width\fi}{!}{%
\begin{tabular}{rrrrr}
\toprule
split time & $F_{ST}$, ours & $F_{ST}$, benchmark & difference & accuracy \\
\midrule
3200 & 0.2405 & 0.2434 & -0.0029 & 0.9983 \\
1600 & 0.1366 & 0.1376 & -0.0010 & 0.9955 \\
800 & 0.0719 & 0.0736 & -0.0017 & 0.9877 \\
400 & 0.0380 & 0.0384 & -0.0004 & 0.9610 \\
200 & 0.0199 & 0.0200 & -0.0001 & 0.8969 \\
100 & 0.0096 & 0.0097 & -0.0001 & 0.7592 \\
50 & 0.0053 & 0.0050 & +0.0003 & 0.6579 \\
25 & 0.0026 & 0.0030 & -0.0004 & 0.5721 \\
\bottomrule
\end{tabular}}
\end{table}

\begin{table}[t]
\centering
\caption{The U-Net depth sweep on the synthetic process, at \unetDepths\ depths and two switch densities, means over \nSeeds\ seeds. ``Receptive field'' is the convolutional span, measured by autograd rather than derived; with statistics pooled along the sequence the same measurement returns \unetWindow\ at every depth, which is the whole sequence. Depth is what sets the receptive field here, and it also sets how many values each normalization statistic is taken over, so this is the one architecture in which the two cannot be varied separately.}
\label{tab:unet}
\resizebox{\ifdim\width>\linewidth\linewidth\else\width\fi}{!}{%
\begin{tabular}{rrrrr}
\toprule
switches/sequence & depth & receptive field & pooled over length & per position \\
\midrule
0.27 & 1 & 22 & 0.9247 & 0.5730 \\
 & 2 & 48 & 0.9252 & 0.6060 \\
 & 3 & 96 & 0.9255 & 0.6493 \\
 & 4 & 192 & 0.9276 & 0.7039 \\
 & 5 & 384 & 0.9326 & 0.7756 \\
 & 6 & 768 & 0.9417 & 0.8482 \\
2.98 & 1 & 22 & 0.6964 & 0.5746 \\
 & 2 & 48 & 0.7066 & 0.6072 \\
 & 3 & 96 & 0.7242 & 0.6492 \\
 & 4 & 192 & 0.7504 & 0.6985 \\
 & 5 & 384 & 0.7808 & 0.7512 \\
 & 6 & 768 & 0.8050 & 0.7928 \\
\bottomrule
\end{tabular}}
\end{table}

\begin{sidewaystable}
\centering
\caption{Normalization configurations of 13 published sequence-labeling models across 4 domains whose labels come in long runs, read from each project's source rather than from its paper. ``Exposed'' means the criterion of Section~\ref{sec:theory} is met: statistics computed from the current input, along the sequence, at inference. Two are. Citations are to each model's publication; the evidence for the row is the code, since the two can disagree (Section~\ref{sec:discussion}). Each row's source link, pinned to the commit that was read, and its provenance are in the released \texttt{exposure\_survey.json}.}
\label{tab:survey}
\resizebox{\ifdim\width>\linewidth\linewidth\else\width\fi}{!}{%
\begin{tabular}{lllc}
\toprule
domain & model & normalization & exposed \\
\midrule
action segmentation & MS-TCN \citep{farha2019mstcn} & none & no \\
action segmentation & ASFormer \citep{yi2021asformer} & InstanceNorm1d, no running stats & \textbf{yes} \\
sleep staging & U-Sleep \citep{perslev2021usleep} & BatchNorm, running stats at eval & no \\
sleep staging & U-Time \citep{perslev2019utime} & BatchNorm, running stats at eval & no \\
sleep staging & DeepSleepNet \citep{supratak2017deepsleepnet} & BatchNorm, moving stats at eval & no \\
sleep staging & TinySleepNet \citep{supratak2020tinysleepnet} & BatchNorm, frozen at eval & no \\
sleep staging & SleepTransformer \citep{phan2022sleeptransformer} & LayerNorm, feature axis & no \\
speaker diarization & SA-EEND \citep{fujita2019eend} & LayerNorm, feature axis & no \\
speaker diarization & EEND-EDA \citep{horiguchi2020eendeda} & LayerNorm, feature axis & no \\
speaker diarization & ECAPA-TDNN \citep{desplanques2020ecapa} & BatchNorm, running stats at eval & no \\
speaker diarization & pyannote.audio SincNet frontend \citep{bredin2020pyannote} & InstanceNorm1d, no running stats & \textbf{yes} \\
genomics & Enformer \citep{avsec2021effective} & BatchNorm, moving averages; LayerNorm, feature axis & no \\
genomics & Basenji \citep{kelley2018sequential} & BatchNorm, moving averages; LayerNorm, feature axis & no \\
\bottomrule
\end{tabular}}
\end{sidewaystable}

\begin{table}[t]
\centering
\caption{Accuracy lost by removing the long-range blocks (those that enlarge the receptive field; Section~\ref{sec:setup}), per divergence level and normalization: ablating them from the trained network against retraining without them. The pooled figures of Section~\ref{sec:ablation} are means over these rows: \ablCostOpen\ against \retCostOpen\ (ratio \ablRatioOpen) with statistics pooled along the sequence, and a ratio of \ablRatioClosed\ per position. Means over \nSeeds\ seeds.}
\label{tab:ablcost}
\resizebox{\ifdim\width>\linewidth\linewidth\else\width\fi}{!}{%
\begin{tabular}{llrrr}
\toprule
normalization & $F_{ST}$ & ablated & retrained & ratio \\
\midrule
pooled over length & 0.2405 & 0.3117 & 0.0351 & 8.9 \\
 & 0.0719 & 0.4273 & 0.0524 & 8.2 \\
 & 0.0380 & 0.4099 & 0.0452 & 9.1 \\
 & 0.0199 & 0.3774 & 0.0467 & 8.1 \\
 & 0.0096 & 0.2289 & 0.0320 & 7.2 \\
per position & 0.2405 & 0.4085 & 0.2754 & 1.5 \\
 & 0.0719 & 0.4491 & 0.3658 & 1.2 \\
 & 0.0380 & 0.4371 & 0.3741 & 1.2 \\
 & 0.0199 & 0.3584 & 0.3199 & 1.1 \\
 & 0.0096 & 0.2084 & 0.2011 & 1.0 \\
\bottomrule
\end{tabular}}
\end{table}

\begin{table}[t]
\centering
\caption{The derived advantage of an oracle given the sequence's class proportion, $1/\sqrt{2\pi n}$, against the value computed exactly. No quantity is fitted. The approximation requires $n \gg 1$ and $p \ll 1$ (Section~\ref{sec:theory}) and fails below $n \approx 1$ because $\bar\pi$ is bounded in $[0,1]$.}
\label{tab:deriv}
\begin{tabular}{rrrr}
\toprule
switches per sequence & computed & $1/\sqrt{2\pi n}$ & error \\
\midrule
0.09 & 0.4760 & 1.3029 & -63\% \\
0.29 & 0.4348 & 0.7471 & -42\% \\
0.99 & 0.3390 & 0.4001 & -15\% \\
2.98 & 0.2239 & 0.2311 & -3\% \\
5.99 & 0.1567 & 0.1631 & -4\% \\
9.89 & 0.1290 & 0.1269 & +2\% \\
19.87 & 0.0858 & 0.0895 & -4\% \\
\bottomrule
\end{tabular}
\end{table}

\end{document}